\documentclass[10pt,twocolumn,letterpaper]{article}

\usepackage{cvpr}
\usepackage{times}
\usepackage{epsfig}
\usepackage{graphicx}
\usepackage{amsmath}
\usepackage{amssymb}
\usepackage{graphicx}
\usepackage{subfigure}
\usepackage{color}

\usepackage[utf8]{inputenc} 
\usepackage[T1]{fontenc}    
\usepackage{url}            
\usepackage{booktabs}       
\usepackage{amsfonts}       
\usepackage{nicefrac}       
\usepackage{microtype}      
\usepackage{times}
\usepackage{helvet}
\usepackage{courier}
\usepackage{graphicx}
\usepackage{subfigure}
\usepackage{algorithm}
\usepackage{algorithmic}
\usepackage{multirow}
\usepackage{booktabs}
\usepackage{authblk}

\usepackage[pagebackref=true,breaklinks=true,letterpaper=true,colorlinks,bookmarks=false]{hyperref}

\cvprfinalcopy 

\def\cvprPaperID{779} 

\ifcvprfinal\fi
\usepackage{caption}
\begin{document}

\title{Unsupervised Cross-dataset Person Re-identification by Transfer Learning of Spatial-Temporal Patterns}
\author[1]{Jianming Lv}
\author[1]{Weihang Chen}
\author[2]{Qing Li}
\author[1]{Can Yang}
\affil[1]{South China University of Technology}
\affil[2]{City University of Hongkongy}
\affil[ ]{\texttt{\{jmlv,cscyang\}@scut.edu.cn,csscut@mail.scut.edu.cn,qing.li@cityu.edu.hk}}
%
\maketitle

\begin{abstract}
  Most of the proposed person re-identification algorithms conduct supervised training and testing on single labeled datasets with small size, so directly deploying these trained models to a  large-scale real-world camera network may lead to poor performance due to underfitting. It is challenging to incrementally optimize the models by using the abundant unlabeled data collected from the target domain.  To address this challenge, we propose an unsupervised incremental learning algorithm, \emph{TFusion}, which is aided by the transfer learning of the pedestrians' spatio-temporal patterns in the target domain. Specifically, the algorithm firstly transfers the visual classifier trained from small labeled source dataset to the unlabeled target dataset so as to learn the pedestrians' spatial-temporal patterns. Secondly, a Bayesian fusion model is proposed to combine the learned spatio-temporal patterns with visual features to achieve a significantly improved classifier. Finally, we propose a learning-to-rank based  mutual promotion procedure to incrementally optimize the classifiers based on the unlabeled data in the target domain. Comprehensive experiments based on multiple real surveillance datasets are conducted, and the results show that our algorithm gains significant improvement compared with the state-of-art cross-dataset unsupervised person re-identification algorithms.

\end{abstract}

\section{Introduction}

\begin{figure}
{
\begin{minipage}[b]{0.5\textwidth}
\includegraphics[width=1\textwidth]{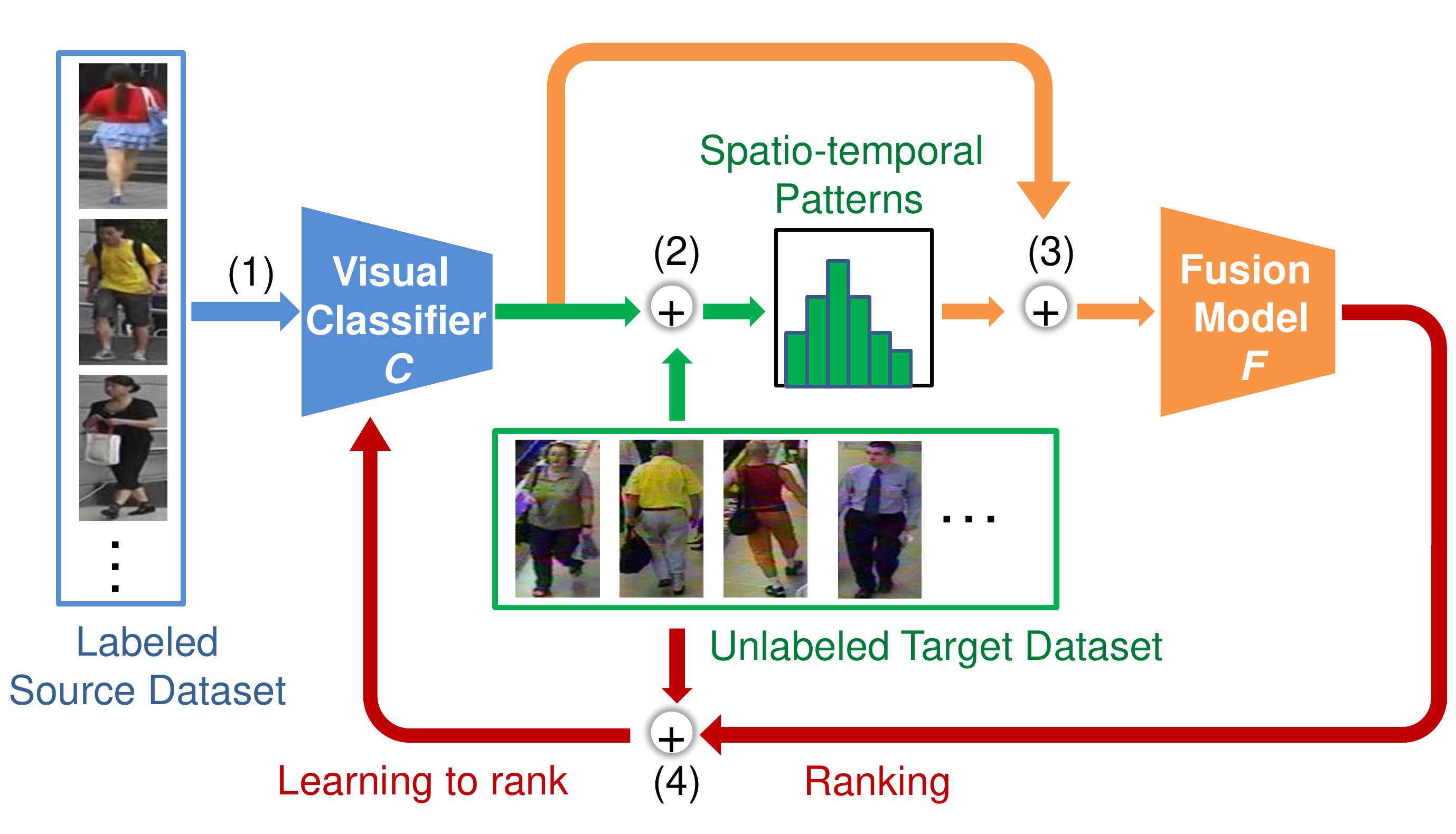}
\end{minipage}
}
\caption{The \emph{TFusion} model consists of 4 steps: (1) Train the visual classifier $\mathcal{C}$ in the labeled source dataset (Section \ref{sec:supervised}); (2) Using $\mathcal{C}$ to learn the pedestrians' spatio-temporal patterns in the unlabeled target dataset (Section \ref{sec:spatio-temporal}); (3) Construct the fusion model $\mathcal{F}$ (Section \ref{sec:fusion}); (4) Incrementally optimize $\mathcal{C}$ by using the ranking results of $\mathcal{F}$ in the unlabeled target dataset (Section \ref{sec:learning-to-rank}).}
\label{fig:model-overview}
\end{figure}

As one of the most challenging and well studied problem in the field of surveillance video analysis, person re-identification (Re-ID) aims to match the image frames which contain the same pedestrian in surveillance videos. The core of these algorithms is to learn the pedestrian features and the similarity measurements, which are view invariant and robust to the change of cameras.

Most of the proposed algorithms \cite{DBLP:conf/cvpr/AhmedJM15}\cite{DBLP:journals/tip/ChenGL16} \cite{DBLP:journals/pr/WuSH17}\cite{DBLP:conf/cvpr/LiaoHZL15} \cite{DBLP:conf/cvpr/Paisitkriangkrai15} \cite{DBLP:journals/tip/TaoGSLYT16} conduct supervised learning on the labeled datasets with small size. Directly deploying these trained models to the real-world environment with large-scale camera networks can lead to poor performance, because the target domain may be significantly different from the  small training dataset. Thus the incremental optimization in real-world deployment is critical to improve the performance of the Re-ID algorithms. However, it is usually expensive and impractical to label the massive online surveillance videos to support supervised learning. How to leverage the abundant unlabeled data is a practical and extremely challenging problem.

To address this problem, some unsupervised algorithms  \cite{DBLP:conf/mm/LiangHHZJX15} \cite{DBLP:journals/pr/MaZGXHLZ17} \cite{DBLP:conf/icip/WangZXG16}  are proposed to extract view invariant features and to measure the similarity of different images in unlabeled datasets. Without powerful supervised tuning and optimization, the performance of above unsupervised algorithms is typically poor. Besides these unsupervised methods applied in a single dataset, a  cross-dataset unsupervised transfer learning algorithm\cite{umdl} is  proposed recently, which transfers the view-invariant representation of a  person's appearance from a source labeled dataset to another unlabeled target dataset by a dictionary learning mechanism, and gains much better performance. However, the performance of the above mentioned algorithms are still much weaker than the supervised learning algorithms. For example, in the CUHK01 \cite{cuhk01} dataset, the unsupervised transfer learning algorithm \cite{umdl} achieves $27.1\%$ rank-1 accuracy, while the accuracy of the  state-of-art supervised algorithm \cite{gog_xqda} can reach to $67\%$ .

In this paper, we propose a novel unsupervised transfer learning algorithm, named \emph{\textbf{TFusion}}, to enable high performance Re-ID in  unlabeled target datasets. Different from the above algorithms which are only based on visual features, we try to learn and integrate with the pedestrians' spatio-temporal patterns in the steps shown in Fig.~\ref{fig:model-overview}. Firstly, we transfer the visual classifier $\mathcal{C}$, which is trained from a small labeled source dataset, to learn the pedestrians' spatio-temporal patterns  in the unlabeled target dataset. Secondly, a Bayesian fusion model is proposed to combine the learned spatio-temporal patterns with visual features to achieve a significantly improved fusion classifier $\mathcal{F}$ for Re-ID in the target dataset. Finally, a learning-to-rank scheme is proposed to further optimize the classifiers based on the unlabeled data. During the iterative optimization procedure, both of the visual classifier $\mathcal{C}$ and the fusion classifier $\mathcal{F}$ are updated in a mutual promotion way.

The comprehensive experiments based on real datasets (VIPeR \cite{viper}, GRID \cite{grid}, CUHK01 \cite{cuhk01} and Market1501 \cite{market1501}) show that \emph{TFusion} outperforms the state-of-art cross-dataset unsupervised transfer algorithm \cite{umdl} by a big margin, and can achieve comparable or even better performance than the state-of-art supervised algorithms using the same datasets.

This paper includes the following contributions:
\begin{itemize}
\item  We present a novel method to learn  pedestrians' spatio-temporal patterns in  unlabeled target datsets by transferring the visual classifier from the source dataset. The algorithm does not require any prior knowledge about the spatial distribution of cameras nor any assumption about how people move in the target environment.

\item We propose a Bayesian fusion model, which  combines the  spatio-temporal patterns learned and the visual features to achieve  high performance of person Re-ID in the unlabeled target datasets.

\item We propose a learning-to-rank based  mutual promotion procedure, which uses the fusion classifier to teach the weaker visual classifier by the ranking results on unlabeled dataset. This mutual learning mechanism can be applied to many domain adaptation problems.
\end{itemize}


\section{Related Work} \label{sec:related-work}

\textbf{Supervised Learning}: Most existing person Re-ID models are supervised, and based on either invariant feature learning \cite{DBLP:conf/eccv/GrayT08} \cite{DBLP:conf/cvpr/LiaoHZL15} \cite{DBLP:conf/cvpr/ZhaoOW14} \cite{DBLP:conf/eccv/YangYYLYL14} ,  metric learning  \cite{DBLP:conf/cvpr/KostingerHWRB12}\cite{DBLP:conf/cvpr/Paisitkriangkrai15} \cite{DBLP:journals/tip/TaoGSLYT16} \cite{DBLP:journals/pami/LisantiMBB15}  or deep learning \cite{DBLP:conf/cvpr/AhmedJM15} \cite{DBLP:journals/tip/ChenGL16} \cite{DBLP:journals/pr/WuSH17} . However, in the practical deployment of Re-ID algorithms in large-scale camera networks, it is usually costly and unpractical to label the massive online surveillance videos to support supervised learning as mentioned in \cite{umdl}.

\textbf{Unsupervised Learning}: In order to improve the effectiveness of the Re-ID algorithms towards large-scale unlabeled datasets, some unsupervised Re-ID methods \cite{DBLP:conf/cvpr/ZhaoOW13}\cite{DBLP:conf/bmvc/WangGX14} \cite{DBLP:conf/mm/LiangHHZJX15} \cite{DBLP:journals/pr/MaZGXHLZ17} \cite{DBLP:conf/icip/WangZXG16}  are proposed to learn cross-view identity-specific information from unlabeled datasets. However, due to the lack of the knowledge about identity labels, these unsupervised approaches usually yield much weaker performance compared to supervised learning approaches.

\textbf{Transfer Learning}: Recently, some cross-dataset transfer learning algorithms\cite{DBLP:journals/tip/MaYT14}  \cite{DBLP:journals/tip/MaLYL15}\cite{umdl}\cite{DBLP:conf/mm/LayneHG13} are proposed to leverage the Re-ID models pre-trained in other labeled datasets to improve the performance  on target dataset. This type of Re-ID algorithms can be classified further into two categories: supervised transfer learning and unsupervised transfer learning  according to whether the label information of target dataset is given or not. Specifically, in the \textbf{supervised transfer learning} algorithms \cite{DBLP:conf/mm/LayneHG13} \cite{DBLP:journals/tip/MaYT14} \cite{DBLP:journals/tip/MaLYL15}, both of the source and target datasets are labeled or have weak labels. \cite{DBLP:conf/mm/LayneHG13}  is based on a SVM multi-kernel learning transfer strategy, and \cite{DBLP:journals/tip/MaLYL15} is based on cross-domain ranking SVMs.  \cite{DBLP:journals/tip/MaYT14}  adopts multi-task metric learning models. On the other hand, the recently proposed cross-dataset \textbf{unsupervised transfer learning} algorithm for Re-ID, UMDL\cite{umdl}, is totally different from above algorithms, and closer to real-world deployment environment where the target dataset is totally unlabeled. UMDL\cite{umdl} transfers the view-invariant representation of a person's appearance from the source labeled dataset to the unlabeled target dataset by dictionary learning mechanisms, and gains much better performance. Although this kind of cross-dataset transfering algorithms are proved to outperform the purely unsupervised algorithms, they still have a long way to catch up the performance of the supervised algorithms, e.g. in the CUHK01\cite{cuhk01} dataset, UMDL \cite{umdl} can achieve $27.1\%$ rank-1 accuracy, while the accuracy of the  state-of-art supervised algorithms \cite{gog_xqda} can reach $67\% $.

Besides the person Re-ID algorithms only based on visual features, some recent research works focus on  \textbf{using the spatio-temporal constraint} in camera networks to improve the Re-ID precision. \cite{DBLP:conf/mmm/HuangHLYWZZ16} considers the distance of cameras and filters the candidates with less possibility.  \cite{DBLP:journals/tcyb/MartinelFM17} models the connection of any pair of cameras by measuring the average similarity score of the images from different cameras, and applies the relationship of cameras to filter the candidates with low probability. \cite{DBLP:journals/cviu/JavedSRS08} makes statistics about the temporal distribution of pedestrians' transferring among different cameras. All of these algorithms  are designed on one single labeled dataset, while our model is adaptive to a cross-dataset transferring learning scenario where the target dataset is totally unlabeled. On the other hand, in above algorithms, the spatio-temporal patterns are learned independent of the visual classifier, and keep fixed at the initialization step. In this paper, we address that the visual classifier and the spatio-temporal patterns can be linked together to conduct an iterative co-train procedure to promote each other.

\section{Preliminaries} \label{sec:preliminaries}
\subsection{Problem Definition of Person Re-ID}

Given a surveillance image containing a target pedestrian, the design goal of a person Re-ID algorithms is to retrieve the surveillance videos for the image frames which contain the same person. For clarity of the problem definition, some notations describing Re-ID are introduced in this section.

Each surveillance image containing a pedestrian is denoted as $S_i$, which is cropped from an image frame of a surveillance video. The time when $S_i$ is taken is denoted by $t_i$, and the ID of the corresponding camera is denoted by $c_i$. The ID of the pedestrian in $S_i$ is denoted as $\Upsilon(S_i)$. Given any surveillance image $S_i$, the person Re-ID problem is to retrieve the images $\{S_j | \Upsilon(S_j) = \Upsilon(S_i)\}$, which contain the same person $\Upsilon(S_i)$.

The traditional strategy of person Re-ID is to train a classifier $\mathcal{C}$ based on visual features to judge whether two given images contain the same person. Given two images $S_i$ and $S_j$, if $\mathcal{C}$ judges that $S_i$ and $S_j$ contain a  same person, it is denoted as $S_i \Vdash_\mathcal{C} S_j$. Otherwise, it is denoted as $S_i \nVdash_\mathcal{C} S_j$.

The false positive error rate of the classifier $\mathcal{C}$ is given by:
\begin{eqnarray}\label{equ:ep}
   E_p &=& Pr(\Upsilon(S_i) \neq  \Upsilon(S_j)| S_i \Vdash_\mathcal{C} S_j)
 \end{eqnarray}
The false negative error rate of $C_s$ is given by:
 \begin{eqnarray}\label{equ:en}
   E_n &=& Pr(\Upsilon(S_i) = \Upsilon(S_j)| S_i \nVdash_\mathcal{C} S_j)
 \end{eqnarray}

\subsection{Cross-Dataset Person Re-ID}

Like most of the traditional person Re-ID algorithms \cite{DBLP:conf/cvpr/LiaoHZL15} \cite{DBLP:conf/cvpr/ZhaoOW14}, we can conduct supervised learning on some public labeled dataset (denoted as $\Omega_s$ below), which is usually of small size, to train a classifier $\mathcal{C}$. While directly deploying the trained $\mathcal{C}$ to a real-world unlabeled target dataset $\Omega_t$ collected from a large-scale camera network, it tends to have poor performance, due to the significant difference between $\Omega_s$ and $\Omega_t$.

How to effectively transfer the classifier trained in a labeled source dataset to another unlabeled target datset is the fundamental challenging problem addressed in this paper.

\section{Model} \label{sec:model}
\subsection{Model overview}

Because most of the time people  move with definite purposes, their trajectories usually follow some non-random patterns, which can be utilized as important clues besides visual features to discriminate different persons. Motivated by this observation, we propose a novel algorithm to transfer the classifier, which is trained in a small source dataset, to learn the spatio-temporal patterns of pedestrians in the unlabeled target dataset. Then we  combine the patterns with the visual features to build a more precise fusion classifier. Furthermore, we adopt a learning-to-rank scheme to incrementally optimize the classifier by using the unlabeled data in the target dataset. The architecture of the model is illustrated in  Fig.~\ref{fig:model-overview}, which contains the following main steps:

\begin {itemize}
\item \textbf{step (1): Supervised Learning in the Labeled Source Dataset}. In this warm-up initialization step, we adopt the supervised learning algorithm such as \cite{zheng2016discriminatively} to learn a visual classier $\mathcal{C}$ from an available small labeled source dataset. In the following steps, further optimization is needed for $\mathcal{C}$ to be applied in a large unlabeled target dataset. (Section \ref{sec:supervised})

\item \textbf{step (2): Transfer Learning of the Spatio-temporal Pattern in the Unlabeled Target Dataset}. In this step, we transfer the classifier $\mathcal{C}$ to the  unlabeled target dataset  to learn pedestrians' spatio-temporal patterns  in the target domain. (Section \ref{sec:spatio-temporal})


\item \textbf{step (3):  Fusion Model for the Target Dataset}. A Bayesian fusion model $\mathcal{F}$ is proposed to combine  the visual classifier $\mathcal{C}$ and the newly learned spatio-temporal patterns  for precise discrimination of pedestrian images. (Section \ref{sec:fusion})

\item \textbf{step (4): Learning-to-rank Scheme for Incremental Optimization of Classifiers}. In this step, we leverage the fusion model $\mathcal{F}$ to further optimize the visual classifier $\mathcal{C}$ based on the learning-to-rank scheme. Firstly, given any surveillance image $S_i$, the fusion model $\mathcal{F}$ is applied to rank the images in the unlabeled target dataset  according to the similarity with $S_i$. Secondly, the ranking results are fed back to incrementally train the visual classifier  $\mathcal{C}$. (Section \ref{sec:learning-to-rank})

\end{itemize}

The model can be iteratively updated by repeating  step $(2)\sim (4)$ until the number of iterations reaches a given threshold or the performance of the classifier converges. In this way, all of the visual classifier $\mathcal{C}$,  the fusion model $\mathcal{F}$, and the spatio-temporal patterns can achieve collaborative optimization.

In the following sections, we will propose the detailed design and analysis of each key component of the model.

\begin{figure}
\centering
{
\begin{minipage}[b]{0.3\textwidth}
\includegraphics[width=1\textwidth]{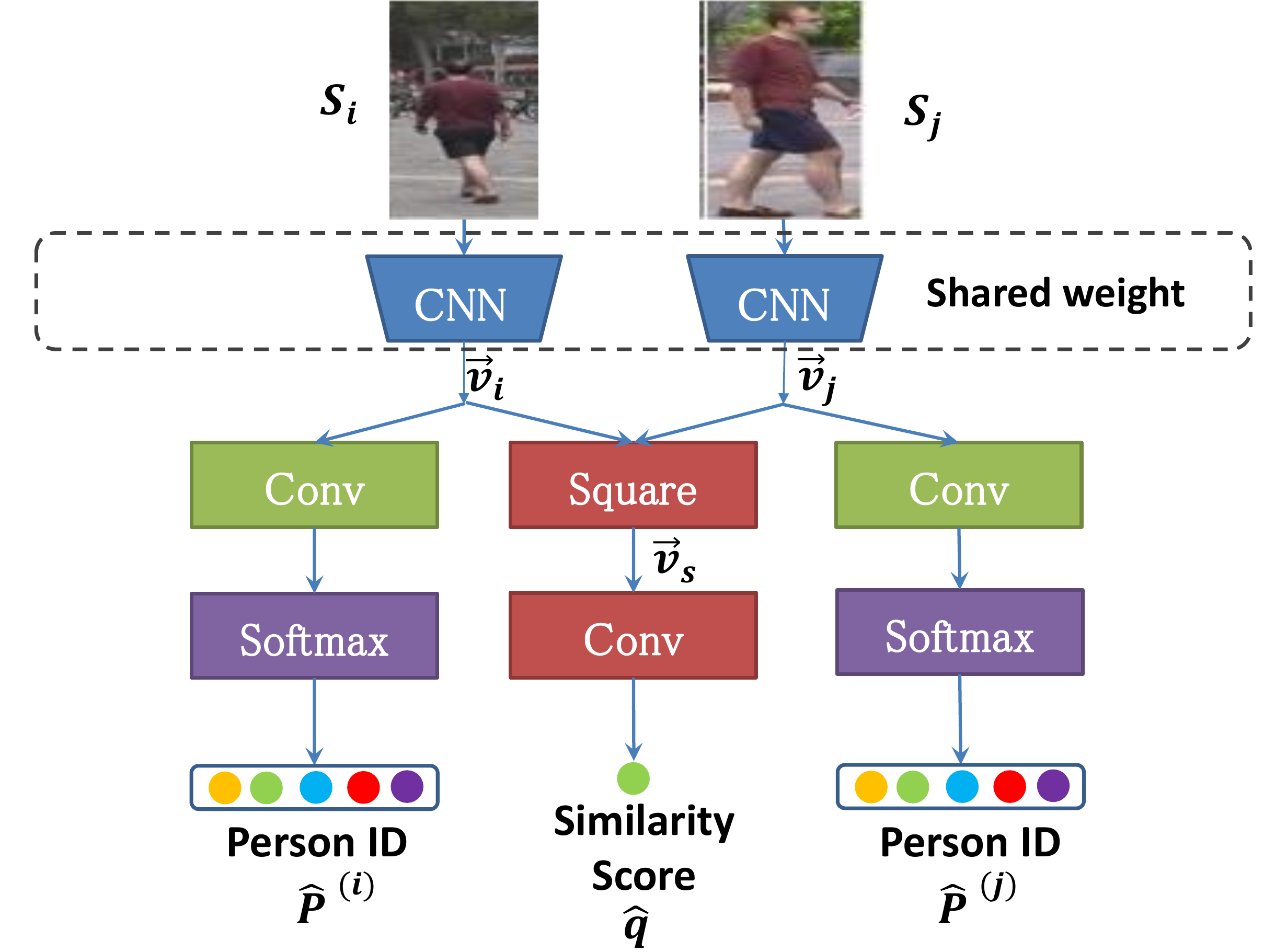}
\end{minipage}
}
\caption{Visual classifier based on CNN.}
\label{fig:visual-classifier}
\end{figure}

\subsection {Supervised Learning in  Labeled Source Dataset} \label{sec:supervised}
As shown in step (1) of Fig.~\ref{fig:model-overview}, the supervised learning is conducted on the labeled source dataset to train the visual classifier $\mathcal{C}$, which measures the matching probability of the given two input images.

We select the recently proposed convolutional siamese network \cite{zheng2016discriminatively} as $\mathcal{C}$, which makes better use of the label information and has good performance in large-scale datasets such as Market1501\cite{market1501}.  The network architecture of $\mathcal{C}$ is shown in Fig.~\ref{fig:visual-classifier}. The network adopts a siamese scheme including two ImageNet pre-trained CNN modules, which share same weight parameters and extract visual features from the input images $S_i$ and $S_j$. The CNN module is achieved from the ResNet-50 network \cite{He2015} by removing its final fully-connected (FC) layer. The outputs of the two CNN modules are flattened into two one-dimensional vectors: $\vec{v_i}$ and $\vec{v_j}$, which act as the embedding visual feature vectors of the input images. Finally, the model predicts the identities ($\hat{P}^{(i)}$ and $\hat{P}^{(j)}$) of the input images, and their similarity score $\hat{q}$. The cross entropy based verification loss and identification loss are adopted for training.  Readers can refer to \cite{zheng2016discriminatively} or our appendix for the detail of the network.

While deploying this classifier to perform Re-ID, given two images $S_i$ and $S_j$ as input, the CNN modules extract their visual feature  vectors $\vec{v_i}$ and $\vec{v_j}$ as shown in Fig.~\ref{fig:visual-classifier}. The matching probability of $S_i$ and $S_j$ is measured as the cosine similarity of the two feature vectors:
\begin {eqnarray} \label{equ:similarity}
Pr(S_i \Vdash_\mathcal{C} S_j| \vec{v_i}, \vec{v_j}) =  \frac{\vec{v_i} \cdot \vec{v_j}}{\parallel \vec{v_i}  \parallel_2 \parallel \vec{v_j}  \parallel_2 }
\end{eqnarray}
If $Pr(S_i \Vdash_\mathcal{C} S_j| \vec{v_i}, \vec{v_j})$  is larger than a predefined threshold constant, $S_i$ and $S_j$ are judged to contain the same person. That is $S_i \Vdash_\mathcal{C} S_j$. Otherwise, they are judged as $S_i \nVdash_\mathcal{C} S_j$.

\subsection{Spatio-temporal Pattern Learning} \label{sec:spatio-temporal}

As reported in \cite{DBLP:journals/cviu/JavedSRS08}, due to the camera network topology, the time interval of pedestrians' transferring among different cameras usually follows specific patterns. These  spatio-temporal patterns  can provide non-visual clues for Re-ID.

Formally, the spatio-temporal pattern of pedestrians' transferring among different cameras can be defined as:
\begin{eqnarray}\label{equ:stpattern}
Pr(\triangle_{ij},c_i,c_j|\Upsilon(S_i) = \Upsilon(S_j)).
\end{eqnarray}
Here $S_i$ is a surveillance image taken at the camera $c_i$ at the time $t_i$, and $S_j$ is another one at the camera $c_j$ at the time $t_j$. $\triangle_{ij} = t_j - t_i$. Eq.(\ref{equ:stpattern}) indicates the probability distribution of the time interval $\triangle_{ij}$ and camera IDs $(c_i,c_j)$ of any pair of image frames $S_i$ and $S_j$ containing  the same person $(\Upsilon(S_i) = \Upsilon(S_j))$.

To calculate the precise value of Eq.(\ref{equ:stpattern}), it is needed to judge whether two images contain the same person firstly. However, this is impossible in unlabeled target datasets where person IDs are unknown. As shown in the step (2) of Fig.~\ref{fig:model-overview}, we propose an approximation solution by transferring the visual classifier $\mathcal{C}$, which is trained in the labeled source dataset, to the unlabeled target dataset. With $\mathcal{C}$, we can make a rough judgment of any pair of images $S_i$ and $S_j$ to achieve the identification result
$S_i \Vdash_\mathcal{C} S_j$ or $S_i \nVdash_\mathcal{C} S_j$. After applying $\mathcal{C}$ to every pair of images in the target dataset, we can obtain the statistics $Pr(\triangle_{ij},c_i,c_j|S_i \Vdash_\mathcal{C} S_j)$,
which indicates the probability distribution of the time interval and camera IDs of any pair of images which seem to contain the same person $(S_i \Vdash_\mathcal{C} S_j)$. On the other hand, we can apply $\mathcal{C}$ to every pair of images in the target dataset to obtain the statistics $Pr(\triangle_{ij},c_i,c_j|S_i \nVdash_\mathcal{C} S_j)$, which indicates the probability distribution of the  time interval and camera IDs of any pair of images which seem to contain different persons. We can infer that:

\begin{footnotesize}
\begin{align} \label{equ:estimate}
   &Pr(\triangle_{ij},c_i,c_j| \Upsilon(S_i) = \Upsilon(S_j)) \nonumber \\
   =& (1-E_n - E_p)^{-1}((1-E_n)*Pr(\triangle_{ij},c_i, c_j|S_i \Vdash_\mathcal{C} S_j) \nonumber \\
   &- E_p*Pr(\triangle_{ij},c_i, c_j|S_i \nVdash_\mathcal{C} S_j))
\end{align}
\end{footnotesize}
Thus, the spatio-temporal pattern $Pr(\triangle_{ij},c_i,c_j| \Upsilon(S_i) = \Upsilon(S_j))$ can be expressed as a function of  $Pr(\triangle_{ij},c_i, c_j|S_i \Vdash_\mathcal{C} S_j)$ and $Pr(\triangle_{ij},c_i, c_j|S_i \nVdash_\mathcal{C} S_j)$, both of which can be measured by the classier $\mathcal{C}$ in the following steps. We first calculate $n$, the number of the image pairs, which satisfy the conditions: 1) they are judged by $\mathcal{C}$ to contain the same person; 2) they are captured at the camera $c_i$ and $c_j$, and 3) the time interval between them is in $[\triangle_{ij} -t, \triangle_{ij} + t]$. Here $t$ is a small threshold.  Then we can use $n/N$ to estimate $Pr(\triangle_{ij},c_i, c_j|S_i \Vdash_\mathcal{C} S_j)$, where $N$ is the total number of  testing image pairs. In a similar way, we can  estimate $Pr(\triangle_{ij},c_i, c_j|S_i \nVdash_\mathcal{C} S_j)$ through counting.

From Eq.(\ref{equ:estimate}), we can infer that while the error rates ($E_p$ and $E_n$) are approaching $0$, the estimated spatio-temporal pattern $Pr(\triangle_{ij},c_i, c_j|S_i \Vdash_\mathcal{C} S_j)$ is approaching the ground-truth pattern $Pr(\triangle_{ij},c_i,c_j| \Upsilon(S_i) = \Upsilon(S_j))$.

\begin{figure*}
{
\begin{minipage}[b]{1\textwidth}
\centering
\includegraphics[width=0.9\textwidth]{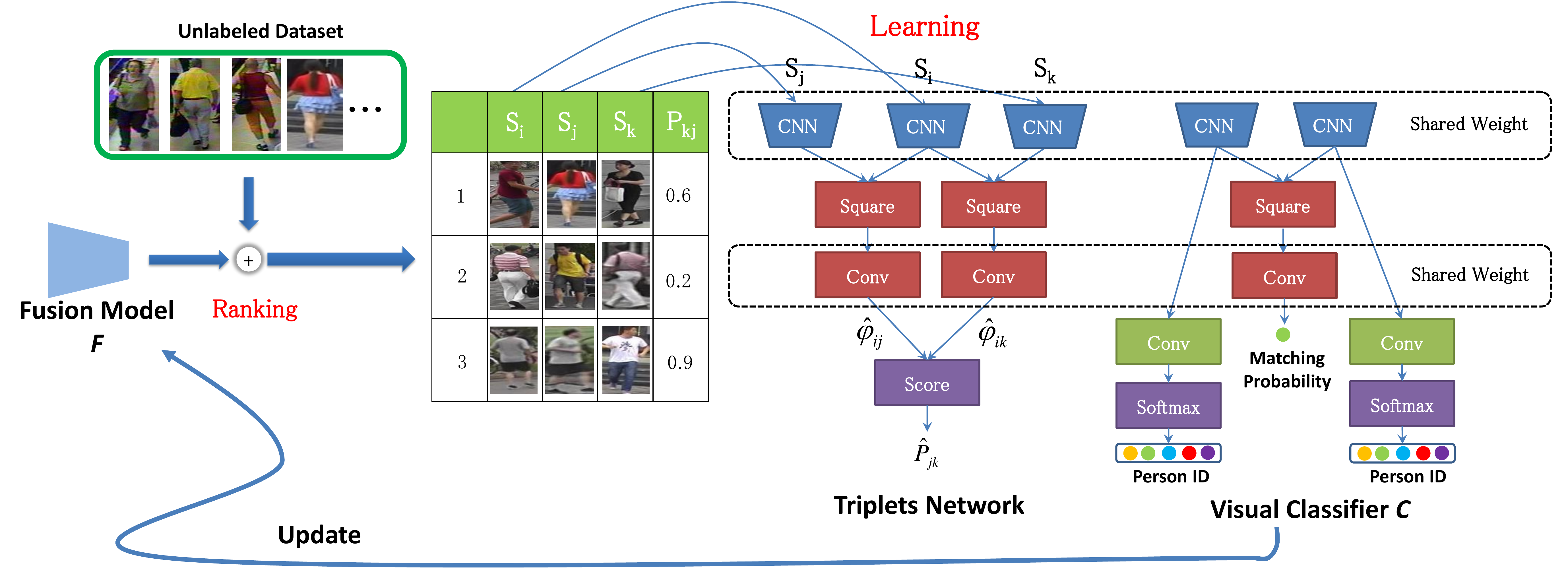}

\end{minipage}
}
\caption{Incremental optimization by the learning-to-rank scheme.}
\label{fig:learning-to-rank}
\end{figure*}

\subsection{Bayesian Fusion model} \label{sec:fusion}

As represented in the last section, the spatio-temporal pattern $Pr(\triangle_{ij},c_i,c_j|\Upsilon(S_i) = \Upsilon(S_j))$, which is  estimated from the visual classifier $\mathcal{C}$, provides a new perspective to discriminate surveillance images besides the visual features used in $\mathcal{C}$. This motivates us to propose a fusion model, which combines the visual features with the spatio-temporal pattern to achieve a composite similarity score of given pair of images, as shown in the step (3) of Fig.~\ref{fig:model-overview}. Formally, the fusion model is based on the conditional probability:

\begin{footnotesize}
\begin{eqnarray} \label{equ:fusion}
Pr(\Upsilon(S_i) = \Upsilon(S_j)|\vec{v_i}, \vec{v_j}, \triangle_{ij},c_i,c_j).
\end{eqnarray}
\end{footnotesize}
Here $S_i$ and $S_j$ are any pair of surveillance images from the target dataset. $S_i$ is taken at the camera $c_i$ at the time $t_i$, and $S_j$ is taken at the camera $c_j$ at the time $t_j$. Their visual feature vectors are denoted as $\vec{v_i}$ and $\vec{v_j}$. The timing interval between them  is $\triangle_{ij} = t_j - t_i$. Eq.(\ref{equ:fusion}) measures the probability of that $S_i$ and $S_j$ contain the same person conditional on their visual features and spatio-temporal information.

According to the Bayesian rule, we have:

\begin{footnotesize}
 \begin{align} \label{equ:bayesian}
   & Pr(\Upsilon(S_i) = \Upsilon(S_j)|\vec{v_i}, \vec{v_j}, \triangle_{ij},c_i,c_j) \nonumber \\
   =& \frac {Pr(\Upsilon(S_i) = \Upsilon(S_j)|\vec{v_i}, \vec{v_j})*Pr(\triangle_{ij},c_i,c_j|\Upsilon(S_i) = \Upsilon(S_j))}{Pr(\triangle_{ij},c_i,c_j)}
\end{align}
\end{footnotesize}
Here $Pr(\Upsilon(S_i) = \Upsilon(S_j)|\vec{v_i}, \vec{v_j})$ indicates the probability of that $S_i$ and $S_j$ contain the same person given their visual features. It can be derived from
$Pr(S_i \Vdash_\mathcal{C} S_j| \vec{v_i}, \vec{v_j})$, which is the matching probability judged by the visual classifier $\mathcal{C}$:

\begin{footnotesize}
\begin{eqnarray} \label{equ:visual}
   &&Pr(\Upsilon(S_i) = \Upsilon(S_j)|\vec{v_i}, \vec{v_j}) \nonumber \\
   &=& Pr(\Upsilon(S_i) = \Upsilon(S_j)| S_i \Vdash_\mathcal{C} S_j)*Pr(S_i \Vdash_\mathcal{C} S_j| \vec{v_i}, \vec{v_j}) +  \nonumber \\
   && Pr(\Upsilon(S_i) = \Upsilon(S_j)| S_i \nVdash_\mathcal{C} S_j)*Pr(S_i \nVdash_\mathcal{C} S_j| \vec{v_i}, \vec{v_j}) \nonumber \\
   &=& (1 - E_p - E_n)* Pr(S_i \Vdash_\mathcal{C} S_j| \vec{v_i}, \vec{v_j}) + E_n
\end{eqnarray}
\end{footnotesize}
On the other hand $Pr(\triangle_{ij},c_i,c_j|\Upsilon(S_i) = \Upsilon(S_j))$ in Eq.~(\ref{equ:bayesian}) indicates the spatio-temporal pattern of pedestrains, and it can be calculated according to Eq.(\ref{equ:estimate}). By substituting Eq.(\ref{equ:estimate}) and (\ref{equ:visual}) into Eq.(\ref{equ:bayesian}), we have:

\begin{footnotesize}
\begin{align} \label{equ:fusion-ex}
   &Pr(\Upsilon(S_i) = \Upsilon(S_j)|\vec{v_i}, \vec{v_j}, \triangle_{ij},c_i,c_j) \nonumber \\
   &=  \frac{(M_1+ \frac{E_n}{1-E_n-E_p})
   ((1-E_n)M_2 - E_pM_3)}{Pr(\triangle_{ij},c_i,c_j)}
\end{align}
\end{footnotesize}
Here, $M_1$,$M_2$, and $M_3$ are defined as follows:

\begin{footnotesize}
\begin{eqnarray} \label{equ:m1m2m3}
   M_1&=& Pr(S_i \Vdash_\mathcal{C} S_j| \vec{v_i}, \vec{v_j}) \nonumber \\
   M_2&=& Pr(\triangle_{ij},c_i, c_j|S_i \Vdash_\mathcal{C} S_j) \nonumber \\
   M_3&=& Pr(\triangle_{ij},c_i, c_j|S_i \nVdash_\mathcal{C} S_j)
\end{eqnarray}
\end{footnotesize}
$M_1$ indicates the judgement of the classifier $\mathcal{C}$ based on the visual features, and it can be measured by $\mathcal{C}$ according to Eq.~(\ref{equ:similarity})). $M_2$ and $M_3$ represent the spatio-temporal patterns of the pedestrians moving in the camera network, and they can be calculated according to the steps mentioned in section \ref{sec:spatio-temporal}. Based on Eq.(\ref{equ:fusion-ex}), we can construct a fusion classifier $\mathcal{F}$, which takes the visual features and spatio-temporal information of two images as input, and outputs their matching probability. As $E_q$ and $E_n$ are unknown in the unlabeled target dataset, Eq.(\ref{equ:fusion-ex}) can not be directly deployed. Thus we substitute $E_q$ and $E_n$ in Eq.~(\ref{equ:fusion-ex}) with two configurable parameters $\alpha$ and $\beta$ to achieve a more general matching probability function of $\mathcal{F}$:

\begin{small}
 \begin{align} \label{equ:fusion-g}
   &Pr(S_i \Vdash_\mathcal{F} S_j |v_i, v_j, \triangle_{ij},c_i,c_j)   \\
   =& \frac{(M_1 + \frac{\alpha}{1 - \alpha - \beta})
   ((1-\alpha)*M_2 - \beta*M_3)}{Pr(\triangle_{ij},c_i,c_j)} (0 \leq \alpha,\beta \leq 1) \nonumber
\end{align}
\end{small}
Here $S_i \Vdash_\mathcal{F} S_j $ means that the classifier $\mathcal{F}$ judges that  $S_i$ and  $S_j$ contain the same person.

In the person re-ID scenario, given any query image $S_i$, we can rank all the images $\{S_j\}$ in the database according to the matching probability $Pr(S_i \Vdash_\mathcal{F} S_j|v_i, v_j, \triangle_{ij},c_i,c_j)$ defined in Eq.(\ref{equ:fusion-g}), and select out the images which have largest probability to contain the same person with $S_i$.

 \subsection{Precision Analysis of the Fusion Model}
In this section, we will analyze the precision of the fusion model  $\mathcal{F}$. Similar with Eq.(\ref{equ:ep}) and (\ref{equ:en}), we define the  false positive error rate $E_p'$ of $\mathcal{F}$ as $Pr(\Upsilon(S_i) \neq  \Upsilon(S_j)| S_i \Vdash_\mathcal{F} S_j)$ and the false negative error rate $E_n'$ of $\mathcal{F}$ as $Pr(\Upsilon(S_i) = \Upsilon(S_j)| S_i \nVdash_\mathcal{F} S_j)$. The following Theorem 1 shows the performance of the fusion model:

\textbf{Theorem 1} \label{theorem-1}: If $E_p + E_n < 1$  and $\alpha + \beta < 1$, we have $E_p'+ E_n' < E_p + E_n$.

Theorem 1 means that the error rate of the fusion model $\mathcal{F}$ may be lower than the original visual classifier $\mathcal{C}$ under the conditions of $E_p + E_n < 1$  and $\alpha + \beta < 1$.  It theoretically shows the effectiveness  to fuse the spatio-temporal patterns with visual features. Due to the page limit, we put the proof of the theorem 1 in the appendix.

\subsection{Incremental Optimization by Learning-to-rank} \label{sec:learning-to-rank}

As shown in Fig.~\ref{fig:model-overview}, the fusion model $\mathcal{F}$ is derived from the visual classifier $\mathcal{C}$ by integrating with spatio-temporal patterns. According to Theorem 1, $\mathcal{F}$ may perform better than $\mathcal{C}$ in the target dataset. That means, given a query image, when using the classifiers to rank the other images according to the matching probability,  the ranking results of $\mathcal{F}$ may be more accurate than that of $\mathcal{C}$. Motivated by this, we propose a novel learning-to-rank based scheme to utilize $\mathcal{F}$ to optimize $\mathcal{C}$ by teaching it with the ranking results in the unlabeled target dataset. Subsequently, the improvement of $\mathcal{C}$ may also derive a better fusion model $\mathcal{F}$. In this mutual promotion procedure, both of the classifiers  $\mathcal{C}$ and $\mathcal{F}$ can get incremental optimization in the unlabeled target dataset.

 The detailed incremental optimization procedure is shown in the Fig.~\ref{fig:learning-to-rank}. In the first step, given any query image $S_i$, the fusion classifier $\mathcal{F}$ is applied to rank the other images in the unlabeled target dataset according to the matching probability defined in Eq.(\ref{equ:fusion-g}). Then we randomly select one image from the top $n (n>0)$ results, and another one  from the results, the rankings of which are in $(n,2n]$.  One of these two images is selected and denoted as $S_j$, and the other one is denoted as $S_k$. The matching probability between $S_i$ and $S_j$ measured by $\mathcal{F}$ is denoted as $\varphi_{i,j}$, and the  matching probability between $S_i$ and $S_k$ is denoted as $\varphi_{i,k}$. The normalized ranking difference between $S_j$ and $S_k$ is defined as: $P_{j,k} = \frac{e^{\varphi_{i,j}- \varphi_{i,k}}}{1+ e^{\varphi_{i,j}- \varphi_{i,k}}}$.

In order to force $\mathcal{C}$ to learn the ranking difference judged by $\mathcal{F}$, we propose a triplets network based on $\mathcal{C}$ to predict the ranking difference. As shown in the Fig.~\ref{fig:learning-to-rank}, the triplets network takes the three images, $S_i$,$S_j$,and $S_k$ as input, and shares the CNN modules with $\mathcal{C}$ to extract visual features. The following square layer and convolutional layer, which are also shared with $\mathcal{C}$, are used to calculate the similarity scores of the image pairs $(S_i, S_j)$ and $(S_i, S_k)$. Their corresponding similarity scores are $\hat{\varphi}_{i,j}$  and $\hat{\varphi}_{i,k}$. In the final score layer, the predicted ranking difference is calculated as $\hat{P}_{j,k} = \frac{e^{\hat{\varphi}_{i,j}-\hat{\varphi}_{i,k}}}{1+ e^{\hat{\varphi}_{i,j}-\hat{\varphi}_{i,k}}}$.

When training the triplets network, the loss function is defined as the cross entropy of the predicted score $\hat{P}_{j,k}$ and the ranking difference $P_{j,k}$ calculated by $\mathcal{F}$: $LOSS_r = -\hat{P}_{j,k}* log(P_{j,k}) - (1-\hat{P}_{j,k})* log(1 - P_{j,k})$.

After training the triplets network, the CNN modules which are shared with the classifier $\mathcal{C}$ get incrementally optimized. In this way, by using the ranking results of the classifier $\mathcal{F}$, we can achieve a upgraded $\mathcal{C}$. Subsequently, the value of $M_1$,$M_2$ and $M_3$  can be updated based on the new $\mathcal{C}$ (Eq.~(\ref{equ:m1m2m3})). With the new $M_1$,$M_2$ and $M_3$,  we can update $\mathcal{F}$, the probability function of which is calculated according to Eq.(\ref{equ:fusion-g}). The mutual promotion of $\mathcal{C}$ and $\mathcal{F}$ can be conducted in multiple iterations to achieve persistent evolving in the unlabeled target dataset, until  the change of the loss $LOSS_r$ among different iterations is less than a threshold.

\begin{table*}[!htbp] \scriptsize
\centering
\caption{Unsupervised transfer learning results.}\label{tab:unsupervised}
\begin{tabular}{l|l||ccc|ccc|ccc|ccc}
\toprule
\multirow{3}*{Source}& \multirow{3}*{Target}& \multicolumn{6}{c|}{Transfer Learning Step} & \multicolumn{6}{|c}{ Incremental Optimization Step} \\
\cline{3-14}
 & & \multicolumn{3}{c|}{Visual Classifier $\mathcal{C}$}& \multicolumn{3}{c|}{Fusion Model $\mathcal{F}$}& \multicolumn{3}{c|}{Visual Classifier $\mathcal{C}$}& \multicolumn{3}{c}{Fusion Model $\mathcal{F}$}\\
\cline{3-14}
 & & rank-1& rank-5& rank-10&rank-1& rank-5& rank-10& rank-1& rank-5& rank-10&rank-1& rank-5& rank-10\\
\hline
\hline
CUHK01& GRID& 10.70 & 20.20 & 23.80 & 30.90 & 63.70 & 79.10 & 17.40 & 33.90 & 41.10 & 50.90 & 78.60 & 88.30\\
VIPeR& GRID& 9.70 & 17.40 & 21.50 & 28.40 & 65.60 & 80.40 & 18.50 & 31.40 & 40.50 & 52.70 & 81.70 & 89.20\\
Market1501& GRID& 17.80 & 31.20 & 36.80& 49.60 & 81.40 & 88.70& 22.30 & 38.10 & 47.20& 60.40 & 87.30 & 93.40\\
\hline
GRID& Market1501& 20.72 & 35.39 & 42.99&51.16& 65.08& 70.04& 22.38&39.25&48.07&58.22& 72.33& 76.84\\
VIPeR& Market1501& 24.70& 40.91& 49.52& 56.18&71.50& 76.48& 25.23& 41.98& 50.33& 59.17& 73.49&78.62\\
CUHK01& Market1501& 29.39& 45.46& 52.55& 56.53& 70.22&74.64& 30.58& 47.09& 54.60& 60.75& 74.44& 79.25\\\bottomrule
\end{tabular}
\end{table*}

\section{Experiment} \label{sec:experiment} 
\subsection{Dataset Setting}


Four widely used benchmark datasets are chosen in our Experiments\footnote{Source Code: https://github.com/ahangchen/TFusion}, including GRID \cite{grid}, Market1501 \cite{market1501}, CUHK01 \cite{cuhk01}, and VIPeR \cite{viper}. As shown in Table.~\ref{tab:unsupervised}, we select one of above datsets as the source dataset and another one as the target dataset to test the performance of  cross-dataset person Re-ID. As mentioned in section \ref{sec:fusion}, the capturing time of each image frame is required to build the fusion model. Thus we choose `Market1501' and `GRID' as target datasets, for they provide the detailed frame numbers in the video sequences, which can be used as timestamps of image frames. The source dataset is chosen without any constraint, because only the image content is used to train the visual classifier $\mathcal{C}$ in the initial step of the model as Fig.~\ref{fig:model-overview}. In this way, there are totally 6 cross-dataset  pairs for experiments as Table.~\ref{tab:unsupervised}.  In each source dataset, all labeled images are used for the pre-training of the visual classifier $\mathcal{C}$. On the other hand, the configurations of the target datasets `Market1501' and `Grid' follow the instructions of these datasets \cite{grid}\cite{market1501} to divide the training and testing set. Specifically, in the `GRID' dataset, a 10-fold cross validation is conducted. In the `Market1501' dataset, 12,936 `bounding-box-train' images are chosen for training and incremental optimization, while 3,368 query images and 19,732 `bounding-box-test' images for single query evaluation.

When adopting Eq.~(\ref{equ:fusion-g}) in the fusion model, $\alpha$ and $\beta$ are two tunable parameters. By default, we set $\alpha=0$ and $\beta=0$.  The performance of different combinations of the parameters are also tested in the following Section~\ref{sec:param}.

\begin{figure}[!htbp]
\centering
\subfigure[]{
\begin{minipage}[b]{0.22\textwidth}
\includegraphics[width=1\textwidth]{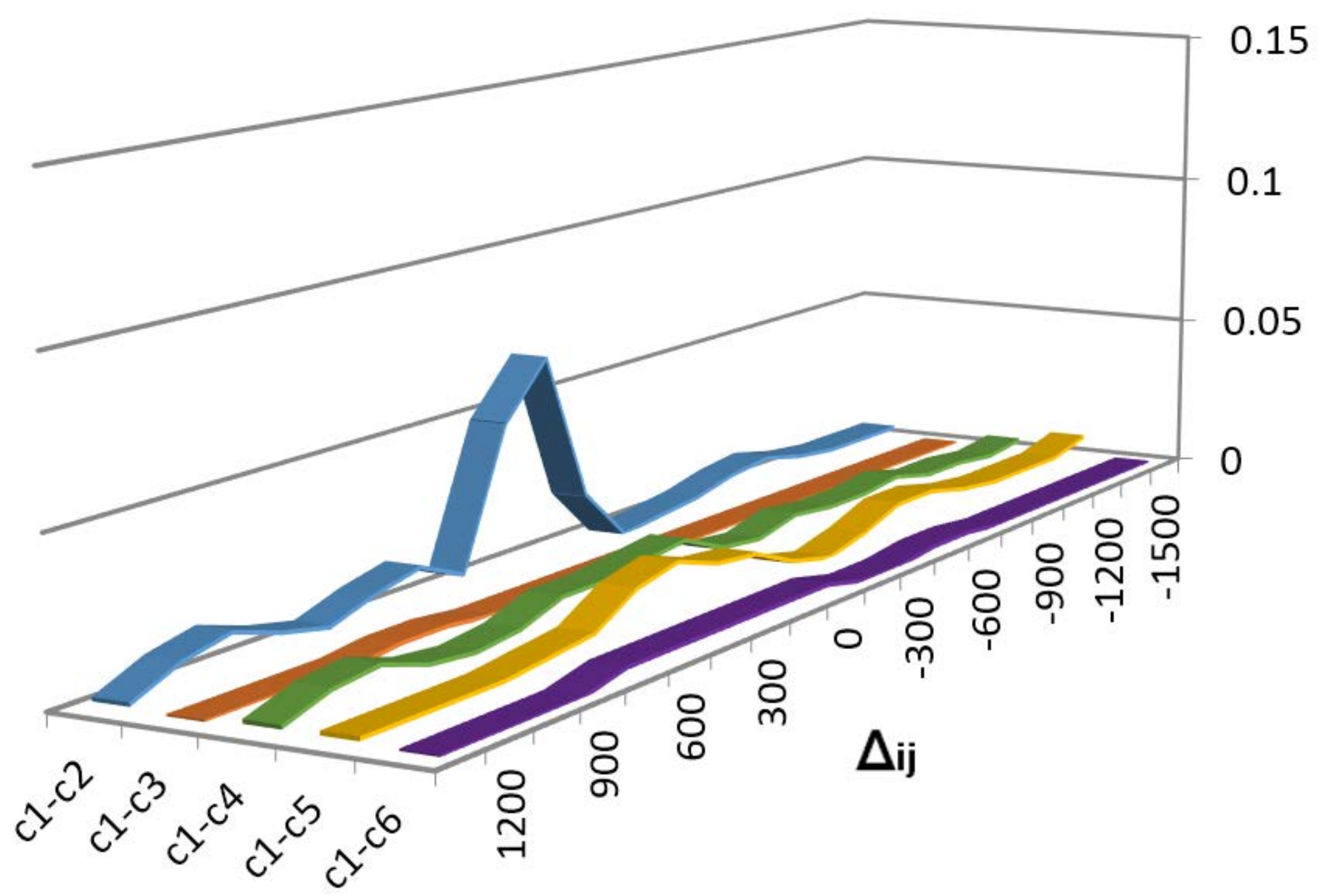}
\end{minipage}
}
\subfigure[]{
\begin{minipage}[b]{0.22\textwidth}
\includegraphics[width=1\textwidth]{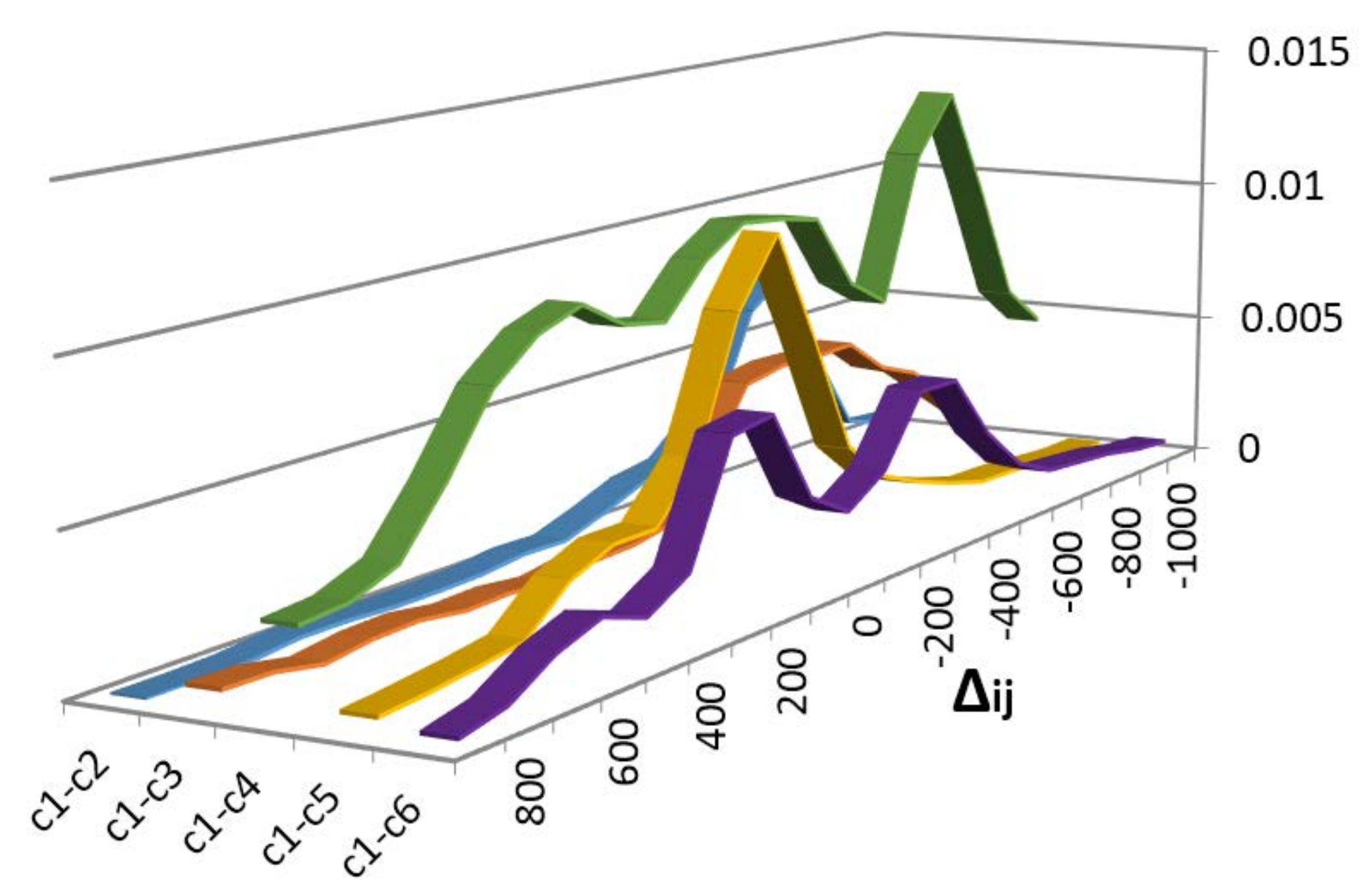}
\end{minipage}
}

\caption{(a)Spatio-temporal pattern in the `GRID' dataset. (b)Spatio-temporal pattern in the `Market1501' dataset.}
\label{fig:delta_distribution}
\end{figure}

\subsection{Learned Spatio-temporal Patterns}

As shown in  Fig.~\ref{fig:model-overview}, learning the spatio-temporal patterns in the unlabeled target dataset is a key step of our fusion model.  As shown in Section \ref{sec:spatio-temporal}, the learned spatio-temporal pattern is represented as the spatio-temporal  distribution $Pr(\triangle_{ij},c_i,c_j|\Upsilon(S_i) \Vdash_\mathcal{C} \Upsilon(S_j))$. Here $S_i$ and $S_j$  are any pair of images captured from the cameras $C_i$ and $C_j$, and they are judged by the visual classifier $\mathcal{C}$ to contain the same person. $\triangle_{ij}$ is defined as : $\triangle_{ij} = t_i - t_j$, where $t_i$ and $t_j$ are the timestamps (frame number) of $S_i$ and $S_j$. Fig.~\ref{fig:delta_distribution} shows the spatio-temporal distribution in the `GRID' and `Market1501' dataset. Due to the limit of  pages, Fig.~\ref{fig:delta_distribution} only shows the distribution related to the first camera in the dataset. The full distribution is attached in the appendix. Fig.~\ref{fig:delta_distribution} shows clearly that the time interval of images from different pairs of cameras follows different non-random distribution, which indicates pedestrians' distinctive temporal patterns to transfer among different locations. This confirms that these spatio-temporal patterns can be used to filter out the matching results with less transferring probability to improve the precision of the person Re-ID system.

\subsection{Re-ID Results}

Table.~\ref{tab:unsupervised} shows the performance of our model in each  training step. Firstly, in the `Transferring Learning Step', the `Visual Classifier $\mathcal{C}$' column means to directly transfer the visual classifier $\mathcal{C}$ trained in the source dataset to the unlabeled target dataset without optimization. Not surprisingly, this kind of  simple transferring method causes  poor performance, due to the variation of data distribution in different datasets. The following `Fusion Model $\mathcal{F}$' column shows that \textbf{the performance of the fusion model, which integrates with the spatio-temporal patterns, gains significant improvement compared with the original visual classifier $\mathcal{C}$}.

The `Incremental Optimization step' in Table.~\ref{tab:unsupervised} means the procedure to use the learning-to-rank scheme to further optimize the model as mentioned in Section \ref{sec:learning-to-rank}. Table.~\ref{tab:unsupervised} shows that,\textbf{ with this incremental learning procedure, the visual classifier $\mathcal{C}$  achieves obvious improvement}. This proves the effectiveness of the learning-to-rank scheme to transfer knowledge from the fusion model $\mathcal{F}$ to the visual classifier $\mathcal{C}$ in the unlabeled target dataset. \textbf{Table.~\ref{tab:unsupervised} also shows that the performance of the fusion model $\mathcal{F}$ achieves significant improvement after the incremental learning}. This is due to the mutual promotion of $\mathcal{F}$ and $\mathcal{C}$ as depicted in Fig.~\ref{fig:model-overview}: a better $\mathcal{C}$ can derive a better $\mathcal{F}$, and a better $\mathcal{F}$ can train the $\mathcal{C}$ into a better one by the learning-to-rank procedure.

~~~~~~~~~~~~~~~~~~~~~

\begin{table}[!htbp]   \scriptsize
\centering
\caption{Compare the precision of TFusion with the state-of-art unsupervised transfer learning methods.}\label{tab:umdl}
\begin{tabular}{l|l|l|ccc}
\toprule
 \multirow{2}*{Method}&\multirow{2}*{Source}&\multirow{2}*{Target}& \multicolumn{3}{c}{Performance}\\
\cline{4-6}
 &&& rank-1& rank-5& rank-10\\
\hline
\hline
\multirow{6}*{UMDL\cite{umdl}}
&Market1501 & GRID & 3.77 & 7.76 &  9.71\\
&CUHK01 & GRID & 3.58 & 7.56 &  9.50\\
&VIPeR & GRID & 3.97 & 8.14 &  10.73\\
\cline{2-6}
&GRID & Market1501 & 30.46 & 45.07 &  52.38\\
&CUHK01 & Market1501 & 29.69 & 44.33 &  51.40\\
&VIPeR & Market1501 & 30.34 & 44.92 &  52.14\\
\hline
\hline
\multirow{6}*{TFusion-uns}
&Market1501& GRID&\multicolumn{1}{|c}{ 60.40} & 87.30 & 93.40\\
&CUHK01& GRID&\multicolumn{1}{|c}{50.90} & 78.60 & 88.30\\
&VIPeR& GRID&\multicolumn{1}{|c}{52.70} & 81.70 & 89.20\\
\cline{2-6}
&GRID& Market1501&\multicolumn{1}{|c}{58.22}& 72.33& 76.84\\
&VIPeR& Market1501&\multicolumn{1}{|c} {59.17}& 73.49&78.62\\
&CUHK01& Market1501& \multicolumn{1}{|c}{ 60.75}& 74.44& 79.25\\

\bottomrule
\end{tabular}
\end{table}

\begin{table}[!htbp] \scriptsize
\centering
\caption{Compare the precision of TFusion with the supervised methods on GRID.}\label{tab:comparison-sup-GRID}
\begin{tabular}{l|ccc}
\toprule
 \multirow{2}*{Method}& \multicolumn{3}{c}{Performance}\\
\cline{2-4}
 & rank-1& rank-5& rank-10\\
\hline
\hline
GOG + XQDA\cite{gog_xqda} & 24.80 &- &58.40\\
HIPHOP+LOMO+CRAFT\cite{HIPHOP_LOMO_CRAFT}& 26.00 &50.60 &62.50\\
SSM\cite{ssm} & 27.20 &- &61.12\\
JLML\cite{jlml}& 37.5 &61.4 &69.4\\
\hline
\hline
TFusion-uns
(Market1501->GRID)&60.40 & 87.30 & 93.40\\
\hline
TFusion-sup & 64.10 & 91.90 & 96.50\\

\bottomrule
\end{tabular}
\end{table}

\begin{table}[!htbp]  \scriptsize
\centering
\caption{Compare the precision of TFusion with the supervised algorithms on Market1501.}\label{tab:comparison-sup-market}
\begin{tabular}{l|cccc}
\toprule
\multirow{2}*{Method}& \multicolumn{4}{c}{Performance}\\
\cline{2-5}
 & rank-1& rank-5 & rank-10 & \\
\hline
\hline
SLSC\cite{cvpr2016spconst} & 51.90& -  & - &\\
LDEHL\cite{nips2016hisembed} & 59.47& 80.73  &86.94 &\\
S-CNN\cite{s-cnn} & 65.88 & - & - & \\
DLCE\cite{zheng2016discriminatively} & 79.51 &90.91 & 94.09 &\\
SVDNet\cite{svdnet} & 82.3 & - & - &\\
JLML\cite{jlml} & 88.8 & - & - & \\
\hline
\hline
TFusion-uns (CUHK01->Market1501)& 60.75& 74.44& 79.25\\
\hline
TFusion-sup & 73.13& 86.43 & 90.46 & \\
\bottomrule
\end{tabular}
\end{table}
We also compare our model, named \textbf{\emph{TFusion}}, with the state-of-art unsupervised cross-dataset person Re-ID algorithm, \emph{UMDL} \cite{umdl}. \emph{UMDL} addresses the similar problem with us, and aims to transfer the visual feature representation from a labeled source dataset to another unlabeled target dataset. UMDL is based on the dictionary learning method and outperforms the state-of-art unsupervised learning algorithms as reported in \cite{umdl}. We compare \emph{TFusion} and \emph{UMDL} under the same dataset configuration and show the results in Table.~\ref{tab:umdl}. \textbf{In all test cases, \emph{TFusion} outperforms \emph{UMDL}  by a large margin.} Especially, for the cases where the target dataset is `GRID', \emph{TFusion} performs extremely well. This may be  attributed to the distinct human motion pattern in the `GRID' dataset, which is collected from a metro station. The fusion with pedestrians' spatio-temporal pattern  can significantly improve the Re-ID performance.

	To observe more clearly the strength of our algorithm to utilize the unlabeled data, we also compare its performance with the state-of-art supervised algorithms deployed on the labeled target datasets. Table.~\ref{tab:comparison-sup-GRID} shows the experimental results in the `GRID' dataset. It is surprising to find that the \emph{TFusion} model, which conducts unsupervised transferring from `Market1501' to `GRID' and does not use the label information of `GRID', \textbf{outperforms the state-of-art supervised algorithms on `GRID'.} This proves again the effectiveness of the fusion with spatio-temporal information. On the other hand, our model can be also run in a supervised mode (denoted as `TFusion-sup' in Table.~\ref{tab:comparison-sup-GRID}),  where both the source dataset and the target dataset are the same.  The performance of \emph{TFusion-sup} is much better than the state-of-art supervised algorithms. It is also interesting to find that the performance of the unsupervised \emph{TFusion} is very close to the supervised version \emph{TFusion-sup}. This shows that the unlabeled data in the target dataset is utilized sufficiently by \emph{TFusion} to achieve good performance. Similarly, Table.~\ref{tab:comparison-sup-market} compares \emph{TFusion} with the state-of-art supervised algorithms on `Market1501'. It also shows that \textbf{our unsupervised transferring model, \emph{TFusion}, can achieve a comparable performance close to the supervised learning models.}

\subsection{Parameter sensitivity} \label{sec:param}

\begin{figure}[!htbp]
  \centering
  \subfigure[]{
  \begin{minipage}[b]{0.2\textwidth}
  \includegraphics[width=1\textwidth]{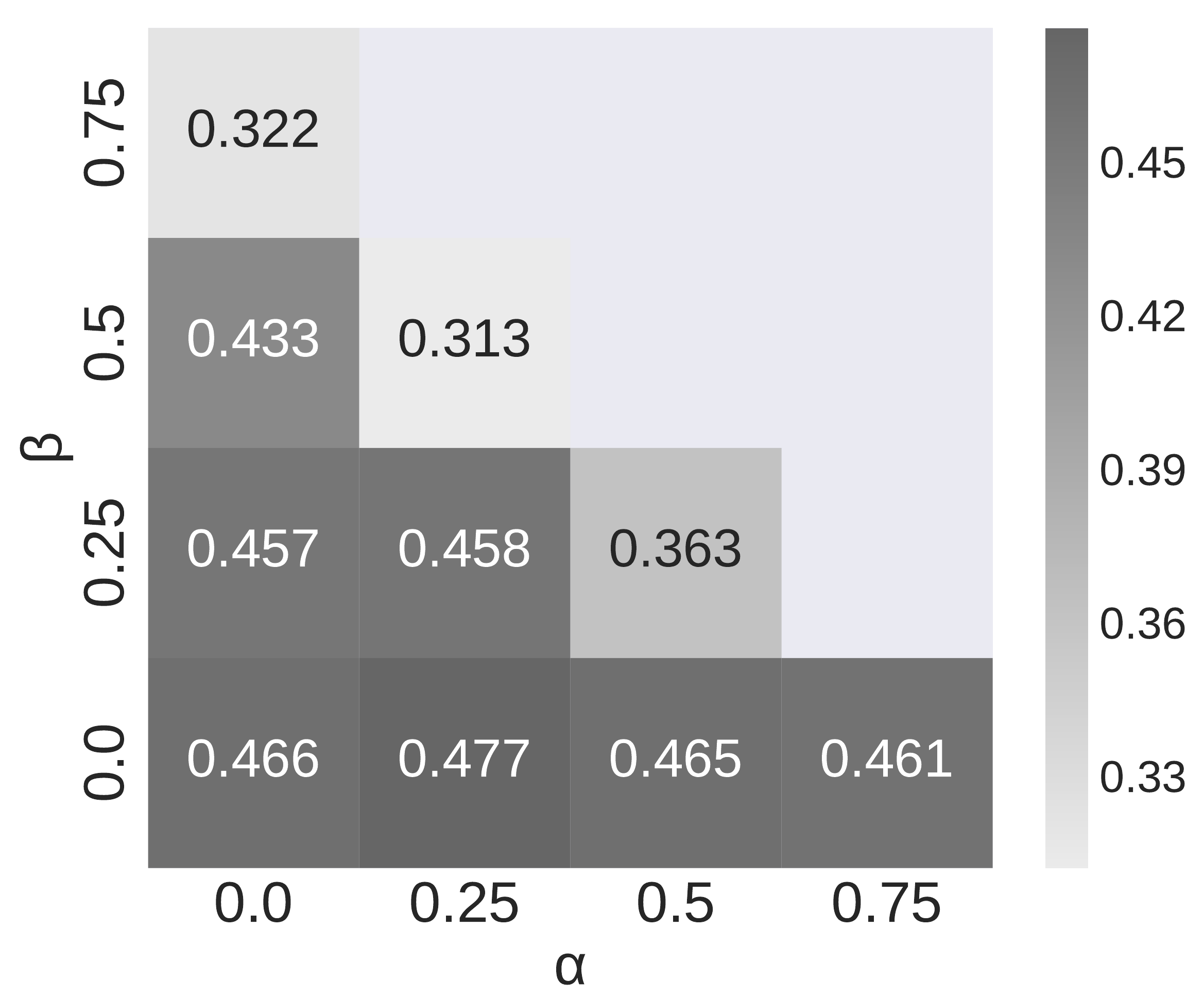}
  \end{minipage}
  \label{fig:GRID_sens}
  }
  \subfigure[]{
  \begin{minipage}[b]{0.2\textwidth}
  \includegraphics[width=1\textwidth]{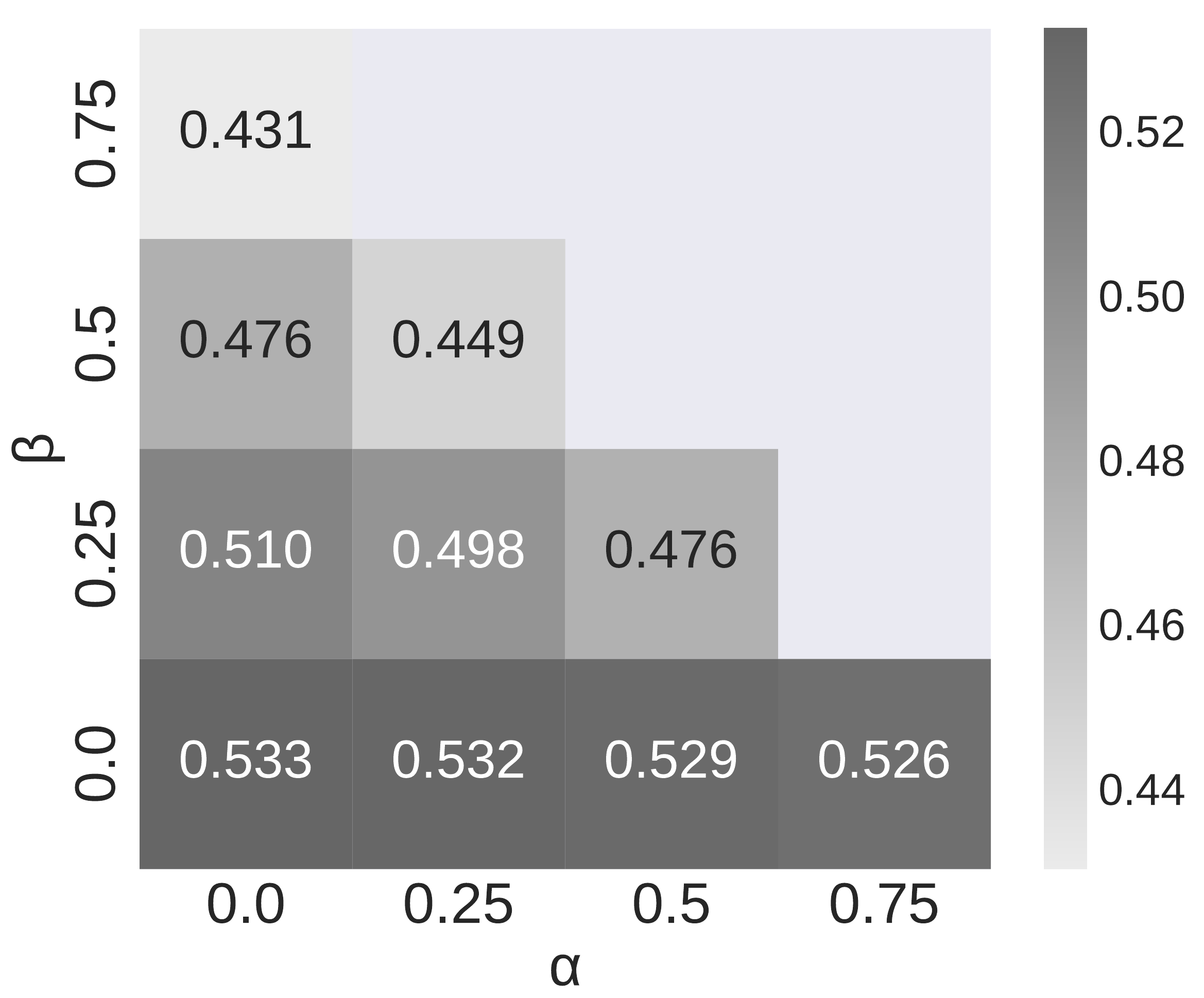}
  \end{minipage}
  \label{fig:market_sense}
  }

  \caption{Performance under different setting of $\alpha$ and $\beta$  (a) in the `Grid' dataset; (b) in  the `Market1501' dataset.}
  \label{fig:real_distribution}
  \end{figure}

As mentioned in Eq.~(\ref{equ:fusion-g}), $\alpha$ and $\beta$ are two tunable parameters in the fusion model. Theorem 1 proves that when $\alpha + \beta < 1$, the fusion model $\mathcal{F}$ may have chance to perform better than the original visual classifier $\mathcal{C}$. Thus, we try different combinations of $\alpha$ and $\beta$, which satisfy $\alpha + \beta < 1$, and test the performance of the fusion model. Fig.~\ref{fig:GRID_sens} shows the rank-1 precision of the models with different $\alpha$ and $\beta$ when transferring from `Market1501' to `GRID', and Fig.~\ref{fig:market_sense} shows the case from `GRID' to  `Market1501'. It shows that the model with smaller $\alpha$ and $\beta$ tends to have better performance.
The combination that $\alpha = 0.25$ and $\beta = 0$ achieves a relatively good performance in both test cases.

\begin{figure}[!htbp]
  \centering
  \subfigure[]{
  \begin{minipage}[b]{0.2\textwidth}
  \includegraphics[width=1\textwidth]{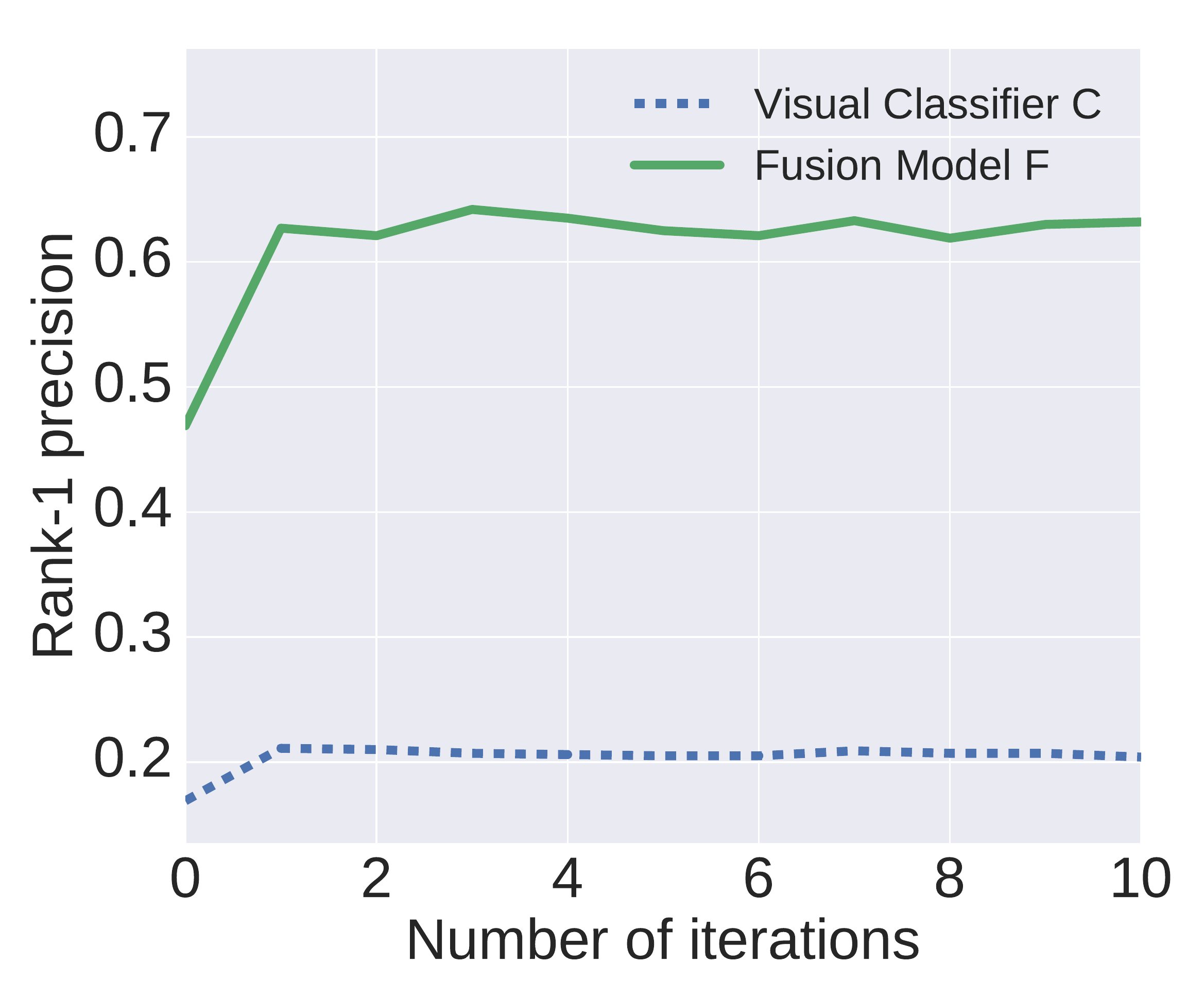}
  \end{minipage}
  \label{fig:market_incremental}
  }
  \subfigure[]{
  \begin{minipage}[b]{0.2\textwidth}
  \includegraphics[width=1\textwidth]{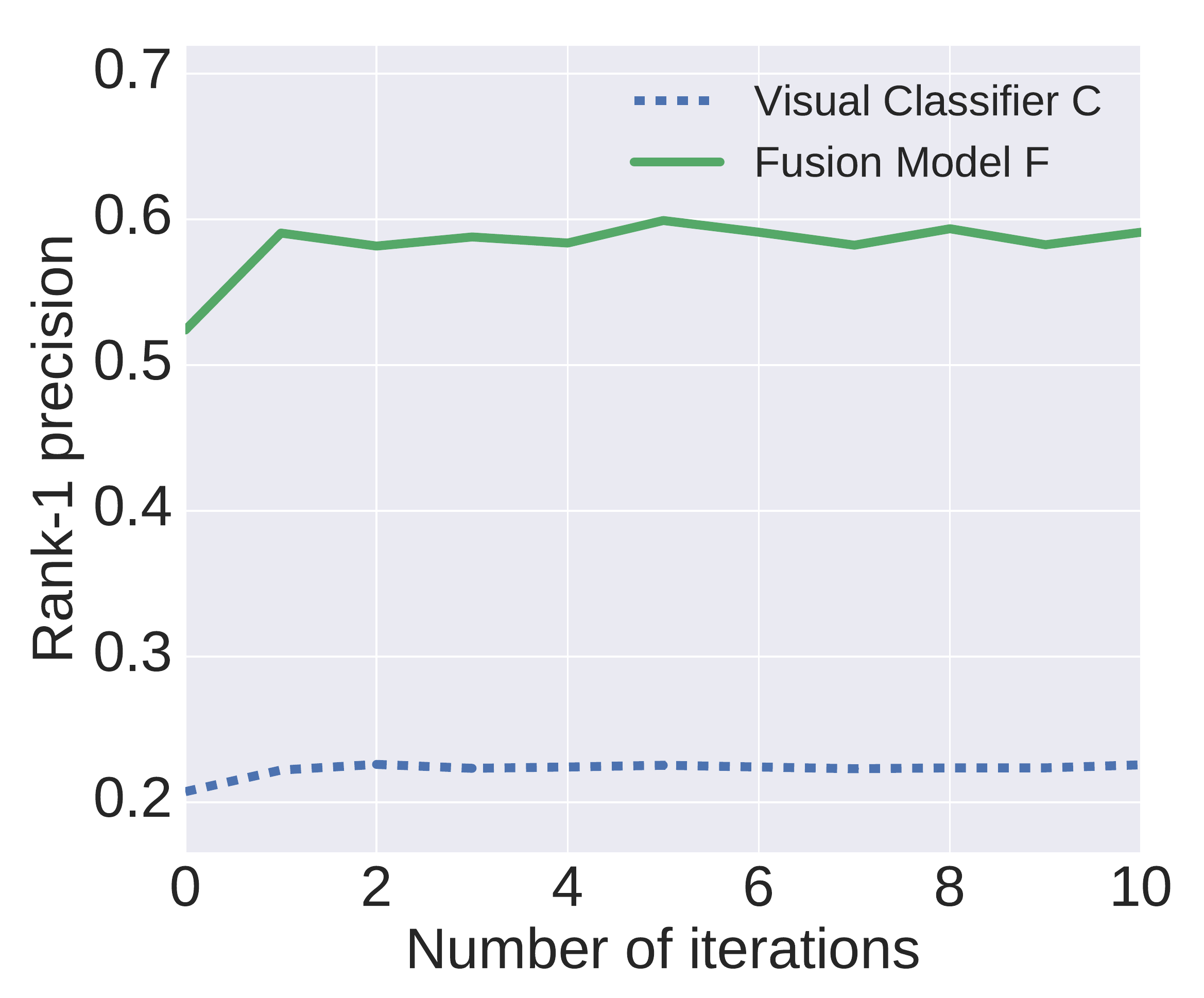}
  \end{minipage}
  }
  \caption{Performance vs. the number of iterations of the learning-to-rank optimization. (a) Performance in the `Grid' dataset. (b) Performance in the `Market1501' dataset.}
  \label{fig:incremental-training}
  \end{figure}

As shown in  Fig.~\ref{fig:learning-to-rank}, the incremental learning procedure consists of iterative learning-to-rank steps.  In each iteration, the fusion model $\mathcal{F}$ is used to train the visual classifier $\mathcal{C}$, and subsequently a more accurate $\mathcal{C}$ can derive a better $\mathcal{F}$. Fig.~\ref{fig:incremental-training} shows how the number of learning-to-rank iterations affects the rank-1 precision of $\mathcal{F}$. It shows that the performance achieves big improvement in the first three iterations, and the precision tends to converge since then. This suggests us that the number of the learning-to-rank iterations can be configured as 3 in the real deployment of \emph{TFusion}.

\section{Conclusions} \label{sec:conclusions}

In this paper, we have presented \emph{TFusion} as a high-performance unsupervised cross-dataset person Re-ID algorithm. In particular, \emph{TFusion} transfers the visual classifier trained in a small labeled source dataset to an unlabeled target dataset by integrating with the spatio-temporal patterns of pedestrians learned in an unsupervised way. Furthermore, an iterative learning-to-rank scheme is proposed to incrementally optimize the model based on the unlabeled data. Experiments show that  \emph{TFusion} outperforms the state-of-art unsupervised cross-dataset transferring algorithm by a big margin, and it also achieves a comparable or even better performance compared with the state-or-art supervised learning algorithms in multiple real datasets.

\textbf{Acknowledgement} The work described in this paper was supported by  the grants from NSFC (No. U1611461), Science and Technology Program of Guangdong Province, China (No. 2016A010101012), and CAS Key Lab of Network Data Science and Technology, Institute of Computing Technology, Chinese Academy of Sciences, 100190, Beijing, China.(No.CASNDST201703).

{\small
\bibliographystyle{ieee}
\bibliography{references}
}

\newpage

\section {Appendix}

\subsection {Architecture of the Visual Classifier $\mathcal{C}$}

\textbf{(Extension of Section 4.2)}

We select the recently proposed convolutional siamese network \cite{zheng2016discriminatively} as $\mathcal{C}$, which makes better use of the label information and has good performance in the large-scale datasets such as Market1501\cite{market1501}. As shown in Fig.~\ref{fig:visual-classifier}, the network adopts a siamese scheme including two ImageNet pre-trained CNN modules, which share the same weight parameters and extract visual features from the input images $S_i$ and $S_j$. The CNN module is achieved from the ResNet-50 network \cite{He2015} by removing its final fully-connected (FC) layer. The outputs of the two CNN modules are flattened into two one-dimensional vectors: $\vec{v_i}$ and $\vec{v_j}$, which act as the embedding visual feature vectors of the input images.

To measure the matching degree of the input images, their feature vectors $\vec{v_i}$ and $\vec{v_j}$ are fed into the following square layer to conduct subtracting and squaring element-wisely: $\vec{v_s} = (\vec{v_i} - \vec{v_j})^2$. Finally, a convolutional layer is used to transform $\vec{v_s}$ into the similarity score as:
\begin{eqnarray}
\hat{q} = sigmoid(\theta_s \circ \vec{v_s})
\end{eqnarray}
.Here $\theta_s$ denotes the parameters in the convolutional layer, $\circ$ denotes the convolutional operation, and $sigmoid$ indicates the $sigmoid$ activation function. By comparing the predicted similarity score with the ground-truth matching result of $S_i$ and $S_j$, we can achieve the \textbf{variation loss} as a cross entropy form:

\begin{footnotesize}
\begin{eqnarray}
LOSS_v = -q \cdot log(\hat{q}) - (1 - q) \cdot log(1-\hat{q})
\end{eqnarray}
\end{footnotesize}
.Here $q = 1$ when $S_i$ and $S_j$ contain the same person. otherwise, $q = 0$.

Besides predicting the similarity score, the model also predicts the identity of each image in the following steps.  Each visual feature vector ($\vec{v_x} (x=i,j)$ ) is  fed into one convolutional layer to be mapped into an one-dimensional vector with the size $K$, where $K$ is equal to the total number of the pedestrians in the dataset. Then the following softmax unit is applied to normalize the output as follows:
\begin{eqnarray}
\hat{P}^{(x)} = softmax ( \theta_x \circ \vec{v_x}) (x=i,j)
\end{eqnarray}
Here $\theta_x$ is the parameter in the convolutional layer and $\circ$ denotes the convolutional operation. The output  $\hat{P}^{(x)}$ is used to predict the identity of the person contained in the input image $S_x(x=i,j)$. By comparing $\hat{P}^{(x)}$ with the ground-truth identify label, we can achieve the \textbf{identification loss} as the cross-entropy form:

\begin{footnotesize}
\begin{eqnarray}
LOSS_{id} = \sum_{k=1}^{K}(-log \hat{P}^{(i)}_k \cdot P^{(i)}_k) + \sum_{k=1}^{K}(-log \hat{P}^{(j)}_k \cdot P^{(j)}_k)
\end{eqnarray}
\end{footnotesize}
Here $P^{(x)} (x=i,j)$ is the identity vector of the input image $S_x$. $P^{(x)}_k = 0$ for all $k$  except $P^{(x)}_t = 1$, where $t$ is ID of the person in the image $S_x$.

\begin{figure}
\centering
{
\begin{minipage}[b]{0.5\textwidth}
\includegraphics[width=1\textwidth]{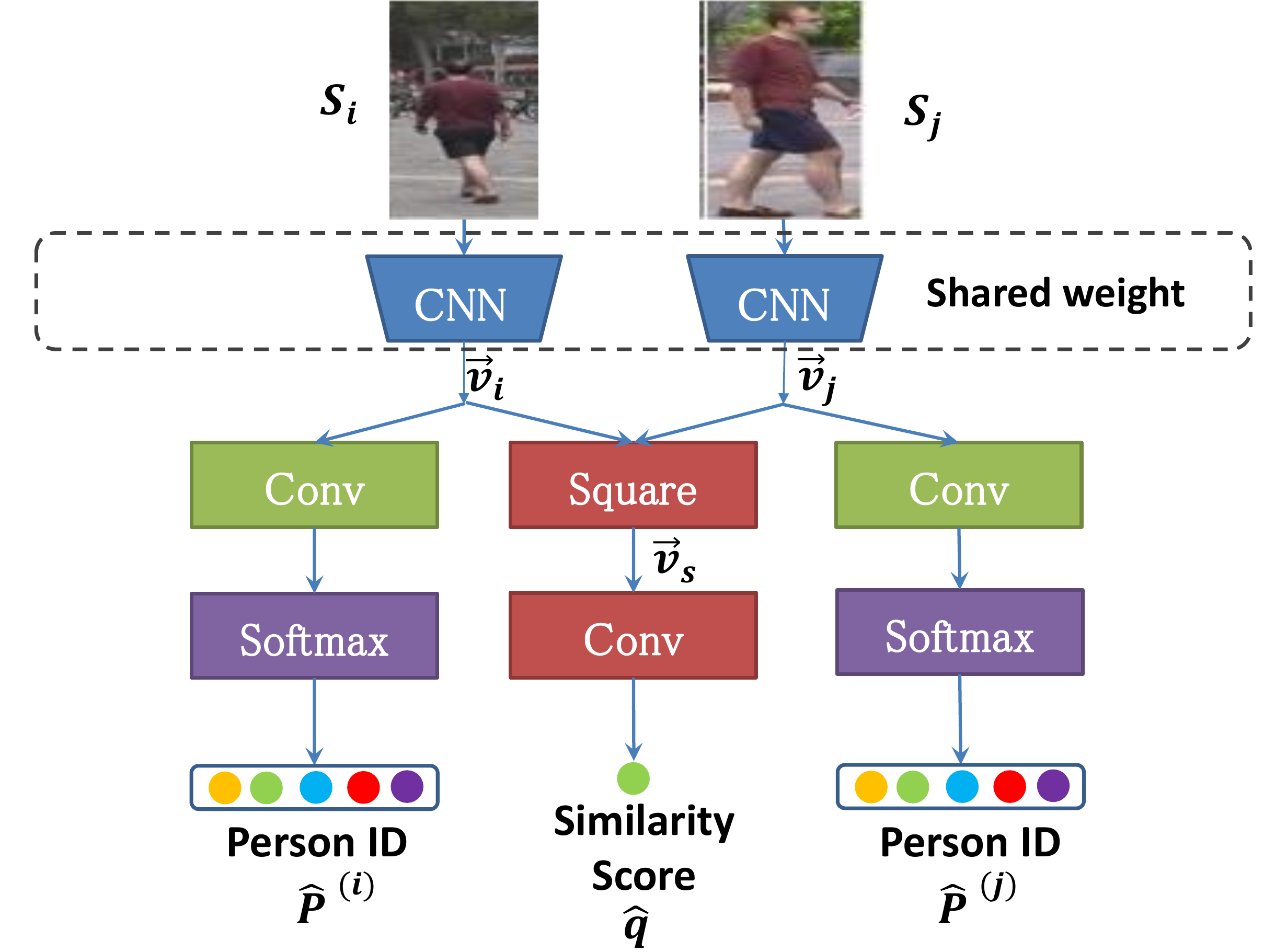}
\end{minipage}
}
\caption{Visual classifier based on CNN.}
\label{fig:visual-classifier}
\end{figure}

The final loss function of the model is defined as:

\begin{footnotesize}
\begin{eqnarray}
LOSS_{all} = LOSS_v + LOSS_{id}
\end{eqnarray}
\end{footnotesize}
According to \cite{jlml}, this kind of  composite loss makes the classifier more efficient to extract the view invariant visual features for Re-ID than the single loss function.

While deploying this classifier to perform Re-ID, given two images $S_i$ and $S_j$ as input, the CNN modules extract their visual feature  vectors $\vec{v_i}$ and $\vec{v_j}$ as shown in Fig.~\ref{fig:visual-classifier}. The matching probability of $S_i$ and $S_j$ is measured as the cosine similarity of the two feature vectors:
\begin {eqnarray} \label{equ:similarity}
Pr(S_i \Vdash_\mathcal{C} S_j| \vec{v_i}, \vec{v_j}) =  \frac{\vec{v_i} \cdot \vec{v_j}}{\parallel \vec{v_i}  \parallel_2 \parallel \vec{v_j}  \parallel_2 }
\end{eqnarray}

\subsection{Proof of Eq.~(5)}

\begin{footnotesize}
\begin{eqnarray}
   &&Pr(\triangle_{ij},c_i,c_j|S_i \Vdash S_j) \nonumber \\
   &=&Pr(\triangle_{ij},c_i, c_j| \Upsilon(S_i) = \Upsilon(S_j))* \nonumber \\
   && Pr(\Upsilon(S_i) = \Upsilon(S_j)|S_i \Vdash S_j) + \nonumber \\
   && Pr(\triangle_{ij},c_i, c_j| \Upsilon(S_i) \neq \Upsilon(S_j))* \nonumber \\
   && Pr(\Upsilon(S_i) \neq \Upsilon(S_j)|S_i \Vdash S_j)\nonumber  \\
   &=& Pr(\triangle_{ij},c_i, c_j| \Upsilon(S_i) = \Upsilon(S_j))*(1 - E_p) +  \nonumber \\
   &&Pr(\triangle_{ij},c_i, c_j| \Upsilon(S_i) \neq \Upsilon(S_j))* E_p
\label{eq:proof-5-1}
\end{eqnarray}
\end{footnotesize}

Similarly, we have:
\begin{footnotesize}
\begin{eqnarray}
   &&Pr(\triangle_{ij},c_i,c_j|S_i \nVdash S_j) \nonumber \\
   &=& Pr(\triangle_{ij},c_i, c_j| \Upsilon(S_i) = \Upsilon(S_j))* \nonumber \\
   && Pr(\Upsilon(S_i) = \Upsilon(S_j)|S_i \nVdash S_j) + \nonumber \\
   && Pr(\triangle_{ij},c_i, c_j| \Upsilon(S_i) \neq \Upsilon(S_j))* \nonumber \\
   && Pr(\Upsilon(S_i) \neq \Upsilon(S_j)|S_i \nVdash S_j)\nonumber  \\
   &=& Pr(\triangle_{ij},c_i, c_j| \Upsilon(S_i) = \Upsilon(S_j))*E_n + \nonumber \\
   && Pr(\triangle_{ij},c_i, c_j| \Upsilon(S_i) \neq \Upsilon(S_j))* (1-E_n)
  \label{eq:proof-5-2}
 \end{eqnarray}
\end{footnotesize}

From (\ref{eq:proof-5-1}) and (\ref{eq:proof-5-2})), we have:
\begin{footnotesize}
\begin{align} \label{equ:estimate}
   &Pr(\triangle_{ij},c_i,c_j| \Upsilon(S_i) = \Upsilon(S_j)) \nonumber \\
   =& (1-E_n - E_p)^{-1}((1-E_n)*Pr(\triangle_{ij},c_i, c_j|S_i \Vdash_\mathcal{C} S_j) \nonumber \\
   &- E_p*Pr(\triangle_{ij},c_i, c_j|S_i \nVdash_\mathcal{C} S_j))
\end{align}
\end{footnotesize}

\rightline{$\Box$}

\subsection{Proof of Theorem 1}
\textbf{Proof of Theorem 1}:
By analyzing the relationship between $Pr(\Upsilon(S_i) = \Upsilon(S_j)|v_i, v_j, \triangle_{ij},c_i,c_j)$  and $ Pr(S_i \Vdash_\mathcal{F} S_j| v_i, v_j, \triangle_{ij},c_i,c_j)$, we have:

\begin{footnotesize}
\begin{align} \label{equ:theorem_1}
   &Pr(\Upsilon(S_i) = \Upsilon(S_j)|v_i, v_j, \triangle_{ij},c_i,c_j) \nonumber \\
   =& Pr(\Upsilon(S_i) = \Upsilon(S_j)| S_i \Vdash_\mathcal{F} S_j)*Pr(S_i \Vdash_\mathcal{F} S_j| v_i, v_j, \triangle_{ij},c_i,c_j) +  \nonumber \\
   & Pr(\Upsilon(S_i) = \Upsilon(S_j)| S_i \nVdash_\mathcal{F} S_j)*Pr(S_i \nVdash_\mathcal{F} S_j| v_i, v_j, \triangle_{ij},c_i,c_j) \nonumber \\
   =& (1- E_p')*Pr(S_i \Vdash_\mathcal{F} S_j| v_i, v_j, \triangle_{ij},c_i,c_j) \nonumber\\
   &+ E_n'*(1- Pr(S_i \Vdash_\mathcal{F} S_j| v_i, v_j, \triangle_{ij},c_i,c_j)) \nonumber \\
   =& (1 - E_p' - E_n')* Pr(S_i \Vdash_\mathcal{F} S_j| v_i, v_j, \triangle_{ij},c_i,c_j) + E_n'
\end{align}
\end{footnotesize}

According to the Eq.(11) of the original paper, we have:

\begin{small}
 \begin{align} \label{equ:fusion-g}
   &Pr(S_i \Vdash_\mathcal{F} S_j |v_i, v_j, \triangle_{ij},c_i,c_j)   \\
   =& \frac{(M_1 + \frac{\alpha}{1 - \alpha - \beta})
   ((1-\alpha)*M_2 - \beta*M_3)}{Pr(\triangle_{ij},c_i,c_j)} (0 \leq \alpha,\beta \leq 1) \nonumber
\end{align}
\end{small}

By substituting Eq.(\ref{equ:fusion-g}) into Eq.(\ref{equ:theorem_1}), we have:

\begin{footnotesize}
\begin{align} \label{equ:theorem_2}
   &Pr(\Upsilon(S_i) = \Upsilon(S_j)|v_i, v_j, \triangle_{ij},c_i,c_j) \nonumber \\
   =& (1 - E_p' - E_n')* \frac{(M_1 + \alpha(1 - \alpha - \beta)^{-1})}{Pr(\triangle_{ij},c_i,c_j)} \nonumber \\
   &* ((1-\alpha) M_2 - \beta M_3)  + E_n'
\end{align}
\end{footnotesize}

On the other hand, from the  Eq.(9) of the original paper, we have:

\begin{footnotesize}
\begin{align} \label{equ:fusion-ex}
   &Pr(\Upsilon(S_i) = \Upsilon(S_j)|\vec{v_i}, \vec{v_j}, \triangle_{ij},c_i,c_j) \nonumber \\
   &=  \frac{(M_1+ \frac{E_n}{1-E_n-E_p})
   ((1-E_n)M_2 - E_pM_3)}{Pr(\triangle_{ij},c_i,c_j)}
\end{align}
\end{footnotesize}

From (\ref{equ:fusion-ex}) and (\ref{equ:theorem_2}) we have:

\begin{footnotesize}
\begin{eqnarray}
 & & (M_1 + E_n(1 - E_p - E_n)^{-1})*
   ((1-E_n)*M_2 - E_p*M_3) \nonumber \\
 &=& (1 - E_p' - E_n')* (M_1 +  \alpha(1 - \alpha - \beta)^{-1}) \nonumber \\
 &&*((1-\alpha) M_2 - \beta M_3)+ E_n'*Pr(\triangle_{ij},c_i,c_j)
\end{eqnarray}
\end{footnotesize}
Thus, we have:

\begin{footnotesize}
\begin{align} \label{equ:theorem_3}
 &\sum\limits_{\triangle_{ij},c_i,c_j} [(M_1 + E_n(1 - E_p - E_n)^{-1}) \nonumber \\
 &*((1-E_n)*M_2 - E_p*M_3)] \nonumber \\
 =& \sum\limits_{\triangle_{ij},c_i,c_j}[((1 - E_p' - E_n')* (M_1 +  \alpha(1 - \alpha - \beta)^{-1}) \nonumber \\
  &* ((1-\alpha) M_2 - \beta M_3)  + E_n'*Pr(\triangle_{ij},c_i,c_j))]
\end{align}
\end{footnotesize}
From (\ref{equ:theorem_3}) have:

\begin{footnotesize}
\begin{eqnarray}\label{equ:theorem_4}
 & & (M_1 + E_n(1 - E_p - E_n)^{-1})
   (1- E_n - E_p) \nonumber \\
 &=& (1 - E_p' - E_n')((1 - \alpha - \beta)M_1 + \alpha)
   p  + E_n'
\end{eqnarray}
\end{footnotesize}
After taking the derivative with respect to  $M_1$ in the both sides of Eq. (\ref{equ:theorem_4}), we can get:

\begin{footnotesize}
\begin{eqnarray} \label{equ:theorem_5}
1 -E_p - E_n = (1- \alpha - \beta)(1-E_p'-E_n')
\end{eqnarray}
\end{footnotesize}
Thus, when $E_p + E_n < 1$ and $\alpha + \beta < 1$, we can infer from Eq.(\ref{equ:theorem_5}) that:

\begin{footnotesize}
\begin{eqnarray}
E_p'+ E_n' < E_p + E_n.
\end{eqnarray}
\end{footnotesize}

\rightline{$\Box$}

\begin{figure}[!htbp]
\centering
\subfigure[]{
\begin{minipage}[b]{0.4\textwidth}
\includegraphics[width=1\textwidth]{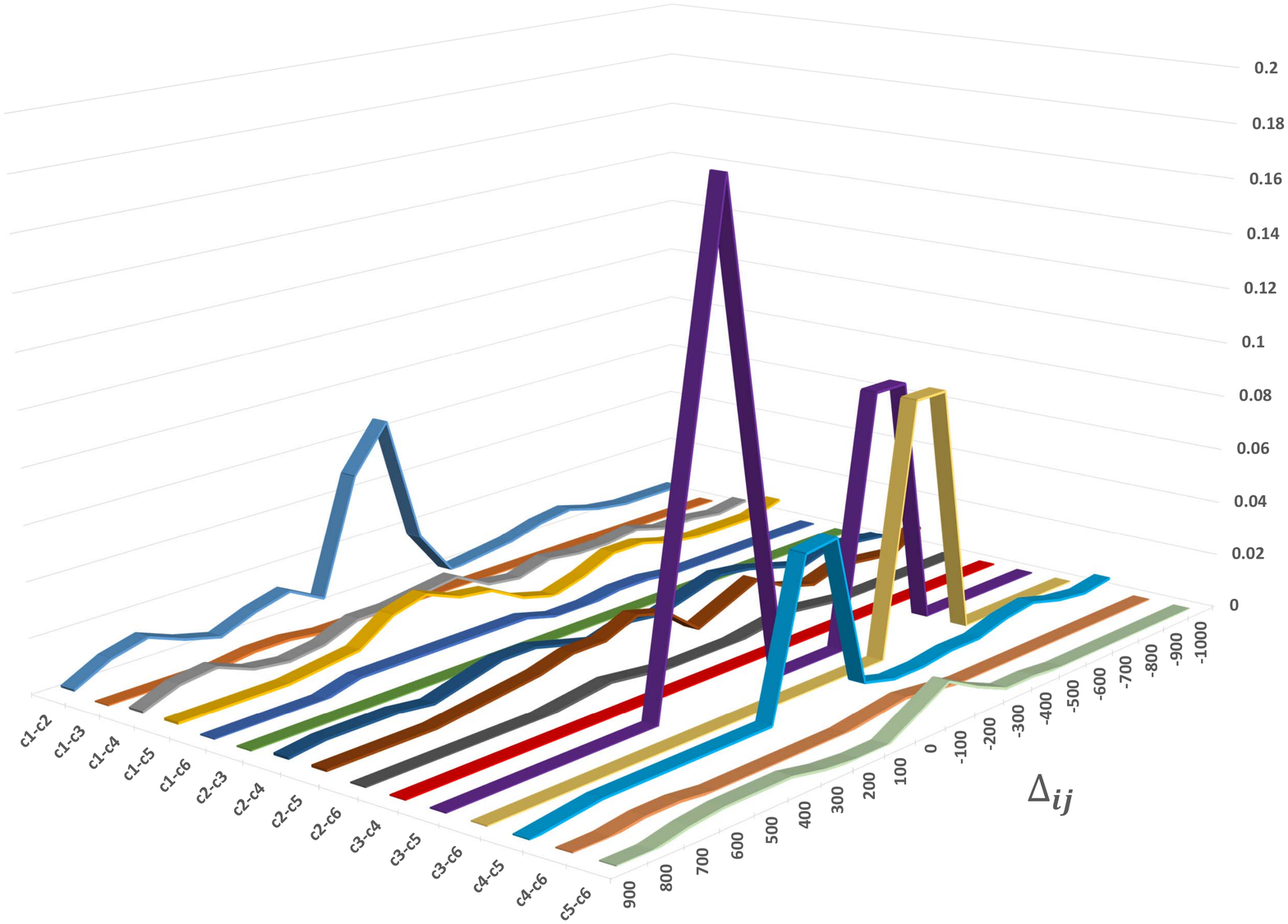}
\end{minipage}
}
\subfigure[]{
\begin{minipage}[b]{0.4\textwidth}
\includegraphics[width=1\textwidth]{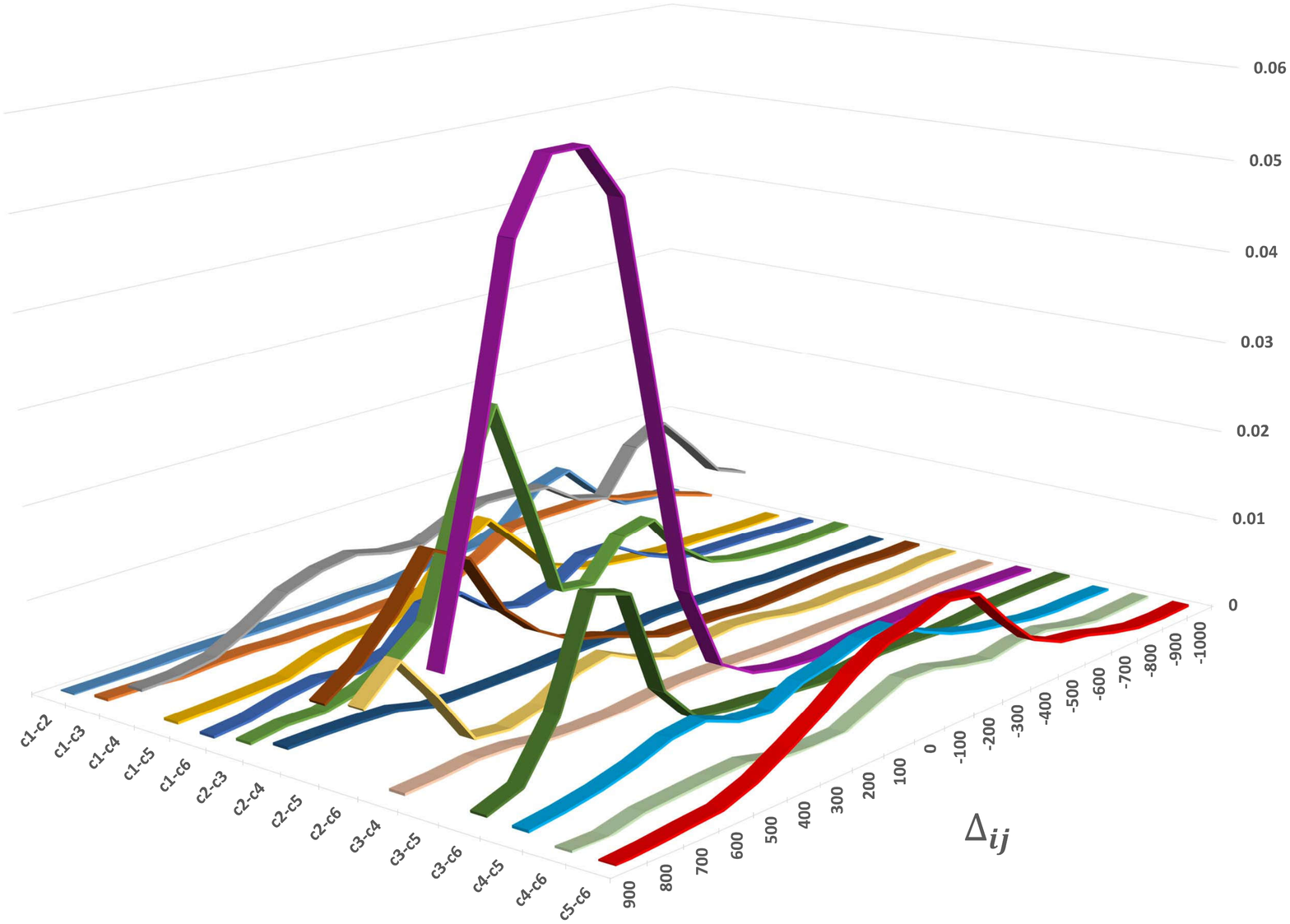}
\end{minipage}

}
\caption{The spatio-temporal distribution $Pr(\triangle_{ij},c_i,c_j|\Upsilon(S_i) \Vdash_\mathcal{C} \Upsilon(S_j))$  learned (a) in the ‘GRID’ dataset, and (b) in the ‘Market1501’ dataset.}
\label{fig:spatio-temporal-dist}
\end{figure}

\subsection{Learned Spatio-temporal Patterns}

\textbf{(Extension of Fig.4)}

Fig.~\ref{fig:spatio-temporal-dist}
shows the spatio-temporal distribution $Pr(\triangle_{ij},c_i,c_j|\Upsilon(S_i) \Vdash_\mathcal{C} \Upsilon(S_j))$ learned in the ‘GRID’ and
‘Market1501’ dataset.

\end{document}